\documentclass[lettersize,journal]{IEEEtran}
\usepackage{amsmath,amsfonts}
\usepackage{algorithmic}
\usepackage{algorithm}
\usepackage{array}
\usepackage[caption=false,font=normalsize,labelfont=sf,textfont=sf]{subfig}
\usepackage{textcomp}
\usepackage{stfloats}
\usepackage{url}
\usepackage{verbatim}
\usepackage{graphicx}
\usepackage{cite}
\usepackage{hyperref}
\usepackage{makecell}
\usepackage{multirow}
\usepackage{colortbl}
\usepackage{rotating}
\usepackage{ragged2e}
\usepackage{tablefootnote}
\usepackage{hhline}
\usepackage[switch]{lineno}

\newtheorem{theorem}{\bf Theorem}
\begin{document}

\title{HOpenCls: Training Hyperspectral Image Open-Set Classifiers in Their Living Environments}

\author{Hengwei~Zhao, Xinyu~Wang, Zhuo~Zheng, Jingtao~Li, Yanfei~Zhong

\thanks{This work was supported by the National Natural Science Foundation of China under Grant No. 42325105.}
\thanks{Hengwei Zhao, Jingtao Li and Yanfei Zhong are with the State Key Laboratory of Information Engineering in Surveying, Mapping and Remote Sensing, Wuhan University, China (e-mail:whu\_zhaohw@whu.edu.cn; jingtaoli@whu.edu.cn; zhongyanfei@whu.edu.cn).}
\thanks{Zhuo Zheng is with the Department of Computer Science, Stanford University, United States (e-mail: zhuozheng@cs.stanford.edu).}
\thanks{Xinyu Wang is with the School of Remote Sensing and Information Engineering, Wuhan University, China (e-mail: wangxinyu@whu.edu.cn).}
}

\markboth{Journal of \LaTeX\ Class Files,~Vol.~14, No.~8, August~2021}%
{Shell \MakeLowercase{\textit{et al.}}: A Sample Article Using IEEEtran.cls for IEEE Journals}


\maketitle

\begin{abstract}
Hyperspectral image (HSI) open-set classification is critical for HSI classification models deployed in real-world environments, where classifiers must simultaneously classify known classes and reject unknown classes. Recent methods utilize auxiliary unknown classes data to improve classification performance. However, the auxiliary unknown classes data is strongly assumed to be completely separable from known classes and requires labor-intensive annotation. To address this limitation, this paper proposes a novel framework, \textit{HOpenCls}, to leverage the unlabeled wild data---that is the mixture of known and unknown classes. Such wild data is abundant and can be collected freely during deploying classifiers in their \textit{living environments}. The key insight is reformulating the open-set HSI classification with unlabeled wild data as a positive-unlabeled (PU) learning problem. Specifically, the multi-label strategy is introduced to bridge the PU learning and open-set HSI classification, and then the proposed gradient contraction and gradient expansion module to make this PU learning problem tractable from the observation of abnormal gradient weights associated with wild data. Extensive experiment results demonstrate that incorporating wild data has the potential to significantly enhance open-set HSI classification in complex real-world scenarios.
\end{abstract}


\section{Introduction}
\IEEEPARstart{H}{yperspectral} image (HSI) can record the spectral characteristics of ground objects~\cite{6555921}. As a key technology of HSI processing, HSI classification is aimed at assigning a unique category label to each pixel based on the spectral and spatial characteristics of this pixel~\cite{FPGA,10078841,10696913,10167502}, which is widely used in agriculture~\cite{WHU-Hi}, forest~\cite{ITreeDet}, city~\cite{WANG2022113058}, ocean~\cite{WHU-Hi} studies and so on.

\begin{figure}[!t]
    \centering
    \includegraphics[width=0.98\columnwidth]{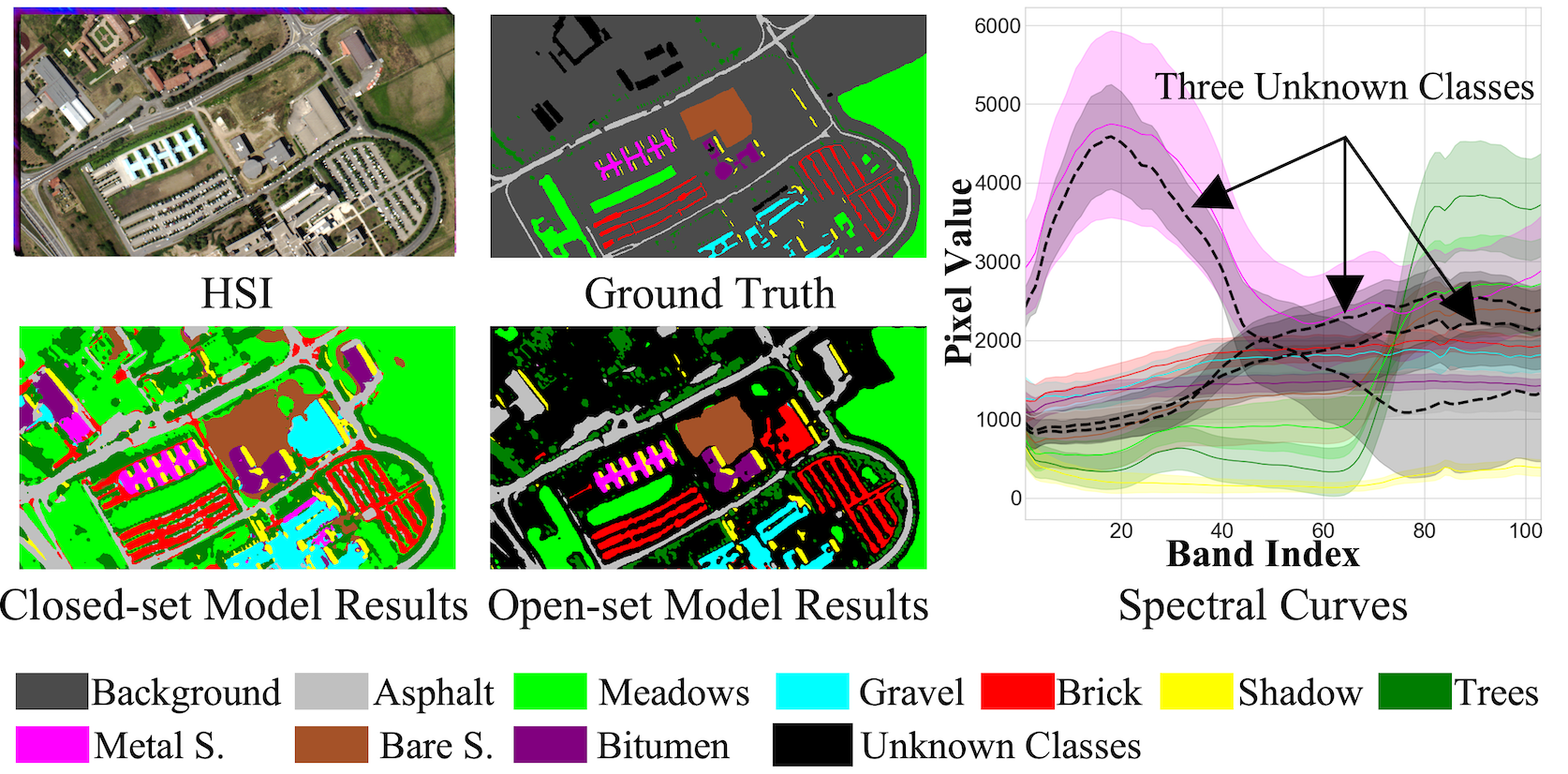}
    \caption{Comparison of classification results between closed-set based classifier and open-set based classifier for the University of Pavia dataset. The dataset originally contains nine \textit{known} land cover classes, however, significant misclassifications occur in the \textit{unknown} classes in closed-set based results. For instance, these unknown buildings with red roofs are misclassified as Bare S., Meadows, and other known materials by closed-set based classifier~\cite{FPGA}. Note that there is a significant overlap in the distribution of spectral curves between known and unknown classes in HSI datasets, which poses a major problem to open-set HSI classification.}
    \label{fig:open_set_example}
\end{figure}

The existing HSI classifiers~\cite{FPGA,10325566,10047983,9573256} typically assume the closed-set setting, where all HSI pixels are presumed to belong to one of the \textit{known} classes. However, due to the practical limitations of field investigations across wide geographical areas and the high annotation costs associated with the limited availability of domain experts, it is inevitable to have outliers in the vast study area~\cite{MDL4OW,Fang_OpenSet,Kang_OpenSet}. These outliers do not belong to any known classes and will be referred as \textit{unknown} classes hereafter. A classifier based on closed-set assumption will misclassify the unknown class as one of the known classes. For example, in the University of Pavia HSI dataset (Fig.~\ref{fig:open_set_example}), objects such as vehicles, buildings with red roofs, carports, and swimming pools are ignored from the original annotations~\cite{MDL4OW}. These objects are misclassified as one of predefined known classes.

Open-set classification (Fig.~\ref{fig:open_set_example}), as a critical task for safely deploying models in real-world scenarios, addresses the above problem by accurately classifying known class samples and rejecting unknown class outliers~\cite{OpenMax,MDL4OW,Fang_OpenSet}. Moreover, the recent advanced researches have explored training with an auxiliary unknown classes dataset to regularize the classifiers to produce lower confidence~\cite{Entropy,WOODS} or higher energies~\cite{Energy} on these unknown classes samples.

Despite its promise, there are some limitations when open-set classification meets HSI. First, the limited number of training samples, combined with significant spectral overlap between known and unknown classes (see Fig.~\ref{fig:open_set_example}), causes the classifier to overfit on the training samples. Second, the distribution of the auxiliary unknown classes dataset may not align well with the distribution of real-world unknown classes, potentially leading to the misclassification of the test-time data. Finally, it is labor-intensive to ensure the collected extra unknown classes dataset does not overlap with the known classes.

To mitigate these limitations, this paper leverages unlabeled ``in-the-wild'' hyperspectral data (referred to as ``wild data''), which can be collected \textit{freely} during deploying HSI classifiers in the open real-world environments, and has been largely neglected for open-set HSI classification purposes. Such data is abundant, has a better match to the test-time distribution than the collected auxiliary unknown classes dataset, and does not require any annotation workloads. Moreover, the information about unknown classes stored in the wild data can be leveraged to promote the rejection of unknown classes in the case of spectral overlap. While leveraging wild data naturally suits open-set HSI classification, it also poses a unique challenge: wild data is not pure and consists of both known and unknown classes. This challenge originates from the marginal distribution of wild data, which can be modeled by the Huber contamination model~\cite{Huber}:
\begin{equation}
    \mathbb{P}_{wild}=\pi\mathbb{P}_{k}+(1-\pi)\mathbb{P}_{u},
    \label{eq:huber_contamination_model}
\end{equation}
where $\mathbb{P}_{k}$ and $\mathbb{P}_{u}$ represent the distributions of known and unknown classes, respectively. Here, $\pi=\pi_{1}+\dots+\pi_{C}$, and $\pi_{c}$ refers to the probability (or class prior~\cite{DistPU}) of the known class $c \in [1,C]$ in $\mathbb{P}_{wild}$.
The known component of wild data acts as noise, potentially disrupting the training process (further analysis can be found in Section~\ref{sec:Methodology}). 

\begin{center}
    \fbox{\begin{minipage}{23em}
        This paper aims to propose a novel framework---\textit{HOpenCls}---to effectively leverage wild data for open-set HSI classification. Wild data is easily available as it's naturally generated during classifier deployment in real-world environments. This framework can be regarded as training open-set HSI classifiers in their \textit{living environments}.
    \end{minipage}}
\end{center}

To handle the lack of ``clean'' unknown classes datasets, the insight of this paper is to formulate a positive-unlabeled (PU) learning problem~\cite{DistPU,T-HOneCls} in the rejection of unknown classes: learning a binary classifier to classify positive (known) and negative (unknown) classes only from positive and unlabeled (wild) data. What's more, the high intra-class variance of positive class and the high class prior of positive class are potential factors that limit the ability of PU learning methods to address unknown class rejection task. To overcome these limitations, the multi-label strategy is introduced to the \textit{HOpenCls} to decouple the original unknown classes rejection task into multiple sub-PU learning tasks, where the $c$-th sub-PU learning task is responsible for classifying the known class $c$ against all other classes. Compared to the original unknown classes rejection task, each sub-PU learning task exhibits reduced intra-class variance and class prior in the positive class.

Beyond the mathematical reformulation, a key contribution of this paper is a novel PU learning method inspired by the abnormal gradient weights found in wild data. First of all, when the auxiliary unknown classes dataset is replaced by the wild data, this paper demonstrates that the adverse effects impeding the rejection of unknown classes originate from the larger gradient weights associated with the component of known classes in the wild data. Therefore, a gradient contraction (Grad-C) module is designed to reduce the gradient weights associated with all training wild data, and then, the gradient weights of wild unknown samples are recovered by the gradient expansion (Grad-E) module to enhance the fitting capability of the classifier. Compared to other PU learning methods~\cite{nnPU,DistPU,PUET,HOneCls}, the combination of Grad-C and Grad-E modules provides the capability to reject unknown classes in a class prior-free manner. Given the spectral overlap characteristics in HSI, estimating class priors for each known class is highly challenging~\cite{T-HOneCls}, and the class prior-free PU learning method is more suitable for open-set HSI classification.

Extensive experiments have been conducted to evaluate the proposed \textit{HOpenCls}. For thorough comparison, two groups of methods are compared: (1) trained with only $\mathbb{P}_{k}$ data, and (2) trained with both $\mathbb{P}_{k}$ data and an additional dataset. The experimental results demonstrate that the proposed framework substantially enhances the classifier's ability to reject unknown classes, leading to a marked improvement in open-set HSI classification performance. Taking the challenging WHU-Hi-HongHu dataset as an example, \textit{HOpenCls} boosts the overall accuracy in open-set classification (Open OA) by 8.20\% compared to the strongest baseline, with significantly improving the metric of unknown classes rejection (F1\textsuperscript{U}) by 38.91\%. The key contributions of this paper can be summarized as follows:
\begin{itemize}
    \item[1)] This paper proposes a novel framework, \textit{HOpenCls}, for open-set HSI classification, designed to effectively leverage wild data. To the best of our knowledge, this paper pioneers the exploration of PU learning for open-set HSI classification.
    \item[2)] The multi-PU head is designed to incorporate the multi-label strategy into \textit{HOpenCls}, decoupling the original unknown classes rejection task into multiple sub-PU learning tasks. As demonstrated in the experimental section, the multi-PU strategy is crucial for bridging PU learning with open-set HSI classification.
    \item[3)] The Grad-C and Grad-E modules, derived from the theoretical analysis of abnormal gradient weights, are proposed for the rejection of unknown classes. The combination of these modules forms a novel class prior-free PU learning method.
    \item[4)] Extensive comparisons and ablations are conducted across: (1) a diverse range of datasets, and (2) varying assumptions about the relationship between the auxiliary dataset distribution and the test-time distribution. The proposed \textit{HOpenCls} achieves state-of-the-art performance, demonstrating significant improvements over existing methods.
\end{itemize}

\section{Related Works}
\subsection{Deep Learning Based HSI Classification}

According to the different learning paradigms, deep learning-based HSI classification approaches can be categorized into patch-based methods and patch-free methods~\cite{FPGA}. Patch-based methods focus on modeling local spectral-spatial information mapping functions:
\begin{equation}
    P_{pb}:R^{S{\times}S}{\rightarrow}R,
    \label{eq:patch_based_framework}
\end{equation}
where a neural network is trained with patches of size $S{\times}S$ extracted from the HSI imagery~\cite{10400402,10050427,9785505}. In contrast, patch-free methods aim to capture global spectral-spatial information mapping functions:
\begin{equation}
    P_{pf}:R^{H{\times}W{\times}C}{\rightarrow}R^{H{\times}W},
    \label{eq:patch_free_framework}
\end{equation}
where $H{\times}W$ represents the spatial size of the HSI imagery and $C$ denotes the number of spectral bands~\cite{8737729,HU2022147,9347487}.
Due to avoiding the utilization of patches, compared with patch-based methods, the GFLOPs of patch-free methods significantly decrease, and the inference time of patch-free methods has significantly improved hundreds of times~\cite{FPGA}.

Most of the existing HSI classifiers are based on the closed-set assumption, which is designed to classify known classes. In contrast, this work focuses on open-set HSI classification, which extends the capability of closed-set based HSI classifiers to not only classify known classes but also reject unknown classes.

\subsection{Open-Set Classification}

The objective of open-set classification is to simultaneously classify known classes and reject unknown classes~\cite{9040673}. Compared to closed-set classification, open-set classification is more challenging due to the incomplete supervision available for rejecting unknown classes~\cite{9857485}. Therefore, this section reviews open-set classification methods from the perspective of unknown classes rejection.

A rich line of method focuses on designing scoring functions for detecting unknown classes, such as the maximum predicted softmax probability (MSP)~\cite{MSP}, OpenMax~\cite{OpenMax}, ODIN score~\cite{ODIN}, Energy score~\cite{Energy}, Entropy score~\cite{Entropy}, MaxLogit score~\cite{KLMatching}, and KL Matching~\cite{KLMatching}. Recent researches have demonstrated that reconstruction loss~\cite{8953952,MDL4OW,Fang_OpenSet} and prototype distance~\cite{Kang_OpenSet,9296325,CACLoss,10415443} can also serve as metrics for rejecting unknown classes. However, these methods are typically trained with known data, and this paper demonstrates that a more robust open-set classifier can be achieved by incorporating naturally occurring wild data from the real-world environments, which can be collected freely.

Another line of research tries to reject unknown classes by using regularization during training~\cite{OE,EOS,Energy,Entropy,WOODS}. These methods typically require an auxiliary unknown classes dataset that is disjoint from $\mathbb{P}_{k}$. For example, models are encouraged to produce lower confidence~\cite{OE,Entropy} or higher energy scores~\cite{Energy} for these auxiliary unknown samples. Similar to this paper, WOODS~\cite{WOODS} also tries to leverage wild data from $\mathbb{P}_{wild}$ by formulating a constrained optimization problem. However, the performance of WOODS is limited by the scarcity of training samples and the significant spectral overlap in HSI.

Different from the abovementioned, this work pioneers the exploration of addressing the open-set HSI classification problem from the perspective of PU learning. Extensive comparisons and ablations demonstrate the clear superiority of the proposed framework.

\subsection{PU Learning}

Early PU learning methods rely on the two-step approach~\cite{FOODY20061,Gong_Wang_Ye_Xu_Lin_2018}, which first extracts reliable negative samples from unlabeled data, and then trains a supervised binary classifier by these positive and selected negative samples. However, the performance of these two-step classifiers is constrained by the reliability of the selected negative samples.

Recent research has shifted towards addressing PU learning using one-step approaches, such as cost-sensitive learning~\cite{9201373,LU2021112584}, label disambiguation~\cite{ijcai2019p590}, and density ratio estimation~\cite{kato2018learning}. What's more, the risk estimation-based methods have proven to be some of the most theoretically and practically effective~\cite{nnPU,ITreeDet,LI2022102947,HOneCls,PUET,DistPU}. However, most of these methods assume that the class prior is known beforehand, which is actually difficult to estimate due to the spectral overlap characteristic in HSI~\cite{T-HOneCls}.

Several researches are striving towards the PU learning without class prior. The generator in the generative adversarial network is replaced by a classifier to learn from PU data in PAN~\cite{PAN}. vPU~\cite{vPU} and T-HOneCls~\cite{T-HOneCls} formalize the PU learning task as the variational problem. P3Mix~\cite{p3mix} proposes a heuristic mixup approach to select partners for positive samples from unlabeled data.

In contrast to these methods, this paper focuses on the challenge of open-set classification. Additionally, in the aspect of class prior-free PU learning, this paper originates from a nontrivial perspective: the abnormal gradient weights associated with wild data.

\section{Methodology}\label{sec:Methodology}
\subsection{Overview}

\begin{figure*}[!t]
    \centering
    \includegraphics[width=0.98\textwidth]{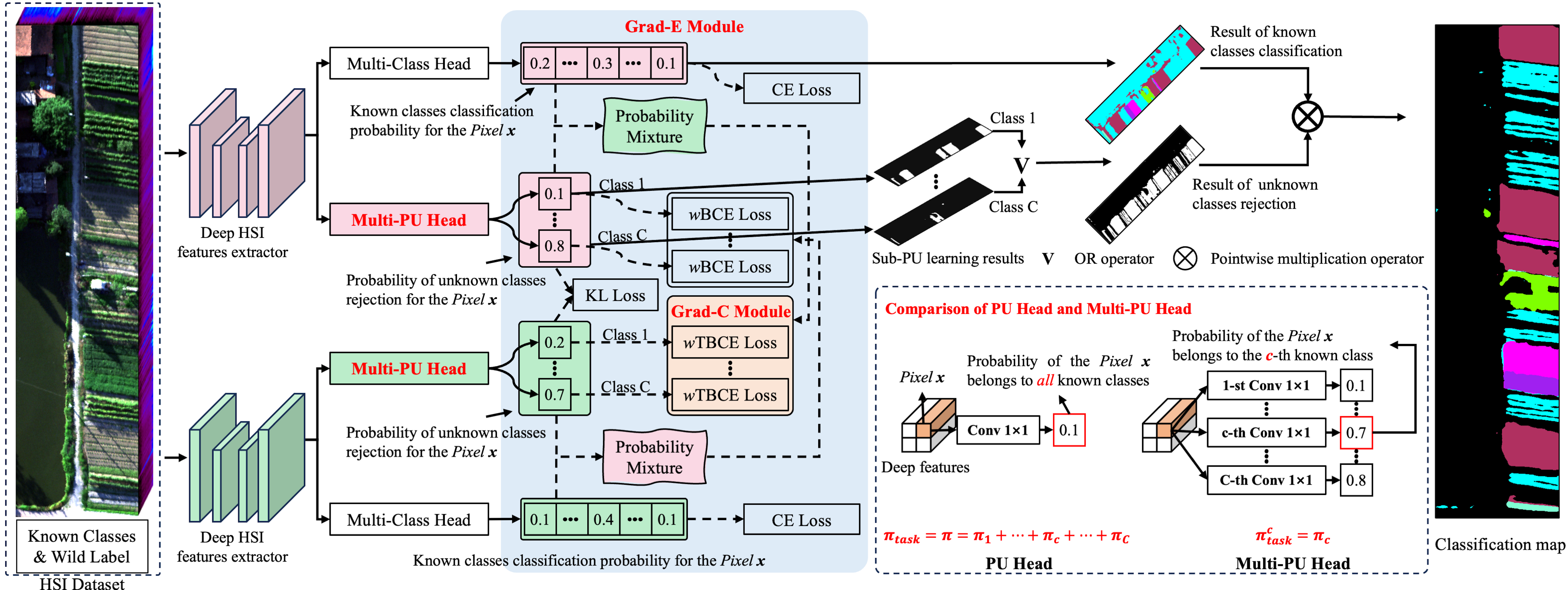}
    \caption{The proposed \emph{HOpenCls} framework. This framework can effectively leverage the wild data for open-set HSI classification. This framework includes multi-PU head, gradient contraction (Grad-C) and gradient expansion (Grad-E) PU learning algorithm. The PU learning component handles the rejection of unknown classes, while the classification of known classes is performed using an existing multi-class HSI classifier.}
    \label{fig:hopencls_framework}
\end{figure*}

The objective of open-set HSI classification is to classify known classes while simultaneously rejecting unknown classes. To achieve this, the proposed \textit{HOpenCls} is designed as a multi-task learning framework, as illustrated in Fig.~\ref{fig:hopencls_framework}. The \textit{HOpenCls} framework is highly flexible and can be implemented by integrating the proposed PU learning classifier with an existing multi-class HSI classifier. The PU learning classifier handles the task of rejecting unknown classes, while the known class classification task is addressed by the existing multi-class HSI classifier.

\noindent \textbf{Problem Settings:}
Let $\mathcal{X}=\{\mathcal{X}_{k}, \mathcal{X}_{u}\} \in \mathbb{R}^{d}$ represent the input space, where $\mathcal{X}_{k}$ and $\mathcal{X}_{u}$ are the input space of known classes and unknown classes, respectively. The output space for the known classes classification task is denoted by $\mathcal{Y}_{k}=\{1,\dots,C\}$, where $C$ is the number of known classes. Additionally, let $\mathcal{Y}_{u}=\{0,1\}$ indicate the output space for the unknown classes rejection task, where 1 corresponds to known classes and 0 corresponds to unknown classes. Suppose the known classes training dataset is denoted as $\mathcal{D}_{k}=\{(\boldsymbol{x}^{i}_{k},y^{i}_{k},y^{i}_{u})\}_{i=1}^{n_{k}}\stackrel{\text{i.i.d}}{\sim}\mathbb{P}_{k}$, where $\boldsymbol{x}^{i}_{k} \in \mathcal{X}_{k}$, $y^{i}_{k} \in \mathcal{Y}_{k}$ is the classification label corresponding to the data $\boldsymbol{x}^{i}_{k}$ of known classes , $y^{i}_{u}=1 \in \mathcal{Y}_{u}$ indicate that $\boldsymbol{x}^{i}_{k}$ belongs to a known class, $n_{k}$ is the number of known classes training samples. The wild training dataset is represented as $\mathcal{D}_{wild}=\{\boldsymbol{x}^{i}_{wild}\}_{i=1}^{n_{wild}}\stackrel{\text{i.i.d}}{\sim}\mathbb{P}_{wild}$, where $\boldsymbol{x}^{i}_{wild} \in \mathcal{X}$, $n_{wild}$ is the number of wild training samples.

Let $q$ and $f$ denote the classifiers for known classes classification and unknown classes rejection, respectively. To reduce computational complexity, a global spectral-spatial feature extractor~\cite{FPGA} is shared by $q$ and $f$. For any test-time data $\boldsymbol{x} \in \mathcal{X}$, the open-set classification result $y$ can be obtained by:
\begin{equation}
    y=Q(\boldsymbol{x}){\otimes}F(\boldsymbol{x}),
\end{equation}
where $\otimes$ is the pointwise multiplication operator. $Q(\boldsymbol{x})$ and $F(\boldsymbol{x})$ are the classification results of $q(\boldsymbol{x})$ and $f(\boldsymbol{x})$, respectively.

\noindent \textbf{Known classes Classification:}
The goal of classifying known classes is to develop a classifier $q$, which can be obtained by minimizing the multi-class classification risk $\mathcal{R}_{k}$:
\begin{equation}
    \mathcal{R}_{k}(q)=\mathbb{E}_{(\boldsymbol{x}_{k},y_{k}){\sim}\mathbb{P}_{k}}\left[\mathcal{L}_{k}(q(\boldsymbol{x}_{k}),y_{k})\right],
    \label{eq:known_classes_risk}
\end{equation}
where $\mathcal{L}_{k}$ is the loss function for multi-class classification. In this paper, for generality, cross entropy (CE) loss ($\mathcal{L}_{ce}$) is used as $\mathcal{L}_{k}$. Eqn.~\ref{eq:known_classes_risk} can be estimated using the empirical averages over the dataset $\mathcal{D}_{k}$:
\begin{equation}
    \hat{\mathcal{R}}_{k}(q)=\frac{1}{n_{k}}\sum_{i=1}^{n_{k}}\mathcal{L}_{ce}(q(\boldsymbol{x}_{k}^{i}),y_{k}^{i}).
    \label{eq:known_classes_average_loss}
\end{equation}

\noindent \textbf{Unknown Classes Rejection:}
The task of rejecting unknown classes involves constructing a binary classifier $f$ that can determine whether a test-time data $\boldsymbol{x}$ belongs to known classes or not. Ideally, the $f$ is obtained by minimizing the unknown classes rejection risk $\mathcal{R}_{u}$:
\begin{equation}
    \mathcal{R}_{u}(f)=\frac{1}{2}\left(\mathcal{R}^{+}_{k}(f)+\mathcal{R}^{-}_{u}(f)\right),
    \label{eq:unknown_classes_risk}
\end{equation}
where $\mathcal{R}^{+}_{k}(f)=\mathbb{E}_{(\boldsymbol{x}_{k},1){\sim}{\mathbb{P}_{k}}}\left[\mathcal{L}_{u}(f(\boldsymbol{x}_{k}),1)\right]$ and $\mathcal{R}^{-}_{u}(f)=\mathbb{E}_{(\boldsymbol{x}_{u},0){\sim}{\mathbb{P}_{u}}}\left[\mathcal{L}_{u}(f(\boldsymbol{x}_{u}),0)\right]$, with $\mathcal{L}_{u}$ representing a binary classification loss function, such as the binary cross entropy (BCE) loss ($\mathcal{L}_{bce}$). $\boldsymbol{x}_{u}$ is the data from unknown classes. However, Eqn.~\ref{eq:unknown_classes_risk} cannot be directly computed in open-set HSI classification due to the absence of training data from unknown classes. Therefore, we focus on designing the multi-PU head, Grad-C module and Grad-E module to reject unknown classes trained by known classes and wild data in the following.

\subsection{Multi-PU Head}

The multi-PU head is designed to introduce the multi-label strategy into the \textit{HOpenCls}. The detailed structure multi-PU head is illustrated in the Fig.~\ref{fig:hopencls_framework}. Based on this design, the original task of unknown classes rejection is decomposed into multiple sub-PU learning tasks: The $c$-th sub-PU learning task is responsible for classifying the $c$-th known class against all other classes. Compared to the original unknown classes rejection task, each sub-PU learning task exhibits reduced intra-class variance of positive class, and the class prior of the $c$-th sub-PU learning task is $\pi_{task}^{c}=\pi_{c}<\pi$, more theoretical analysis about class prior can be found in Theorem ~\ref{theorem}.

In the $c$-th sub-PU learning task, the risk is defined as $\mathcal{R}_{pu}^{c}(f^{c})$:
\begin{equation}
    \mathcal{R}_{pu}^{c}(f^c)=\frac{1}{2}\left({\mathcal{R}^{c}_{k}}^{+}(f^c)+{\mathcal{R}_{wild}}^{-}(f^c)\right),
    \label{eq:cth_class_unknown_classes_risk}
\end{equation}
where ${\mathcal{R}^{c}_{k}}^{+}(f^c)=\mathbb{E}_{(\boldsymbol{x},1){\sim}{\mathbb{P}_{k}^{c}}}\left[\mathcal{L}_{u}(f^{c}(\boldsymbol{x}),1)\right]$, ${\mathcal{R}_{wild}}^{-}(f^c)=\mathbb{E}_{(\boldsymbol{x},0){\sim}{\mathbb{P}_{wild}}}\left[\mathcal{L}_{u}(f^{c}(\boldsymbol{x}),0)\right]$. Here, $\mathbb{P}_{k}^{c}$ represents the margin distribution of the $c$-th known class. The function $f^{c}$ is the binary classifier for the $c$-th sub-PU learning task.

The Eqn.~\ref{eq:cth_class_unknown_classes_risk} can be estimated using the empirical averages over the dataset $\mathcal{D}_{k}$ and the wild dataset $\mathcal{D}_{wild}$:
\begin{equation}
    \begin{aligned}
        \hat{\mathcal{R}}_{pu}^{c}(f^c)=&\frac{1}{2}(\frac{1}{n_{k}^{c}}\sum_{i=1}^{n_{k}^{c}}\mathcal{L}_{u}(f^{c}(\boldsymbol{x}^{ci}_{k}),1)+\\&\frac{1}{n_{wild}}\sum_{i=1}^{n_{wild}}\mathcal{L}_{u}(f^{c}(\boldsymbol{x}^{i}_{wild}),0)),
    \end{aligned}
    \label{eq:cth_class_unknown_classes_average_loss}
\end{equation}
where $\boldsymbol{x}^{ci}_{k}$ is the training sample of known class $c$ from the $\mathcal{D}_{k}$ and $n_{k}^{c}$ is the number of training data of known class $c$. The overall risk for unknown classes rejection with multi-PU head can be estimated as follows:
\begin{equation}
    \hat{\mathcal{R}}_{mpu}(f)=\sum_{c=1}^{C}\hat{\mathcal{R}}_{pu}^{c}(f^{c}).
    \label{eq:unknown_classes_risk_multi_pu_head_average_loss}
\end{equation}

As illustrated in Fig.~\ref{fig:hopencls_framework}, a sample $\boldsymbol{x}$ will be classified as belonging to known classes if any of the sub-PU learning classifiers identify it as known:
\begin{equation}
    F(\boldsymbol{x})=F^{1}(\boldsymbol{x}) \vee \dots \vee F^{c}(\boldsymbol{x}) \vee \dots \vee F^{C}(\boldsymbol{x}),
    \label{eq:unknown_rejection_multi_pu_head}
\end{equation}
where $F^{c}(\boldsymbol{x})$ is the classification result of $f^{c}(\boldsymbol{x})$, the symbol $\vee$ represents the logical OR operation.

\subsection{Grad-C and Grad-E Modules for PU Learning}

\begin{figure}[!t]
    \centering
    \includegraphics[width=0.98\columnwidth]{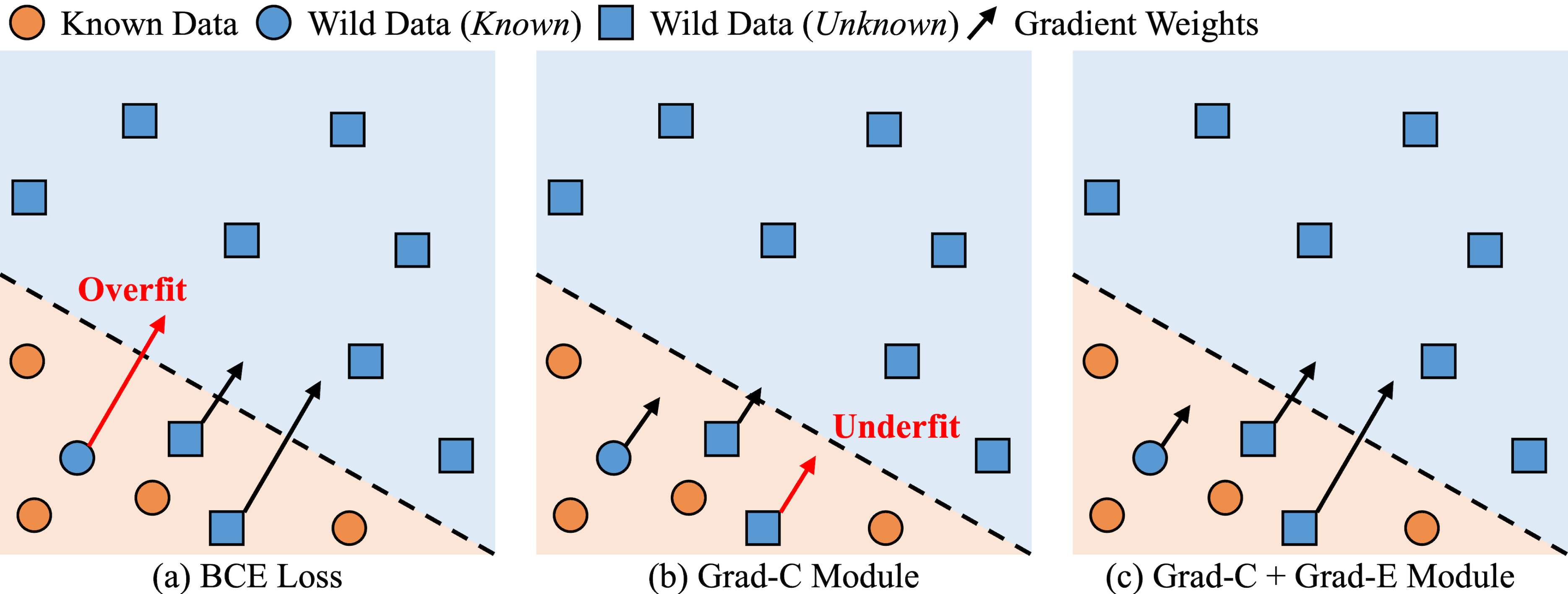}
    \caption{Comparison of different modules against to wild known data. (a) The negative impact of unknown classes data replaced by wild data stems from the larger gradient weights associated with wild known data. (b) The Grad-C module reduces the gradient weights associated with both wild known and unknown data. (c) The Grad-E module restores the gradient weights for the wild unknown data by weighting mechanism.}
    \label{fig:grad}
\end{figure}

The Grad-C and Grad-E modules are designed to reject unknown classes from the perspective of PU learning. First, this paper reveals that the negative impact of replacing a pure unknown dataset by wild data stems from the larger gradient weights associated with wild data. Then, the Grad-C and Grad-E modules are designed to mitigate the adverse effects of these abnormal gradient weights. The comparison of different modules in the aspect of gradient weights is shown in Fig.~\ref{fig:grad}.

\noindent \textbf{Gradient Analysis:}
A commonly used choice for $\mathcal{L}_{u}$ in binary classification is the $\mathcal{L}_{bce}$:
\begin{equation}\nonumber
    \mathcal{L}_{bce}(f(\boldsymbol{x}),y_{u})=-y_{u}\log(f(\boldsymbol{x}))-(1-y_{u})\log(1-f(\boldsymbol{x})).
    \label{eq:binary_cross_entropy}
\end{equation}
The gradient of the $\mathcal{L}_{bce}$ with respect to the parameters $\boldsymbol{\theta}$ of the binary classifier $f$ is given by:
\begin{equation}
    \frac{\partial \mathcal{L}_{bce}(f(\boldsymbol{x}),y_{u})}{\partial \boldsymbol{\theta}}=(\frac{1-y_{u}}{1-f(\boldsymbol{x})}-\frac{y_{u}}{f(\boldsymbol{x})})\nabla_{\boldsymbol{\theta}}f({\boldsymbol{x}}).
    \label{eq:gradient_binary_cross_entropy}
\end{equation}

When the unknown data ($\boldsymbol{x}_{u},0$) is replaced with wild data ($\boldsymbol{x}_{wild},0$), a wild known data will be assigned a larger gradient weight (Eqn.~\ref{eq:gradient_wild_bce}) if it is predicted as known classes with high confidence ($f(\boldsymbol{x}_{wild}) \rightarrow 1$). The $f$ will overfit all wild known data as unknown classes, which disrupts the training process for rejecting unknown classes, as shown in Fig.~\ref{fig:grad}..
\begin{equation}
    \frac{\partial \mathcal{L}_{bce}(f(\boldsymbol{x}_{wild}),0)}{\partial \boldsymbol{\theta}}={\frac{1}{1-f(\boldsymbol{x}_{wild})}}\nabla_{\boldsymbol{\theta}}f(\boldsymbol{x}_{wild}).
\label{eq:gradient_wild_bce}
\end{equation}

\noindent \textbf{Grad-C Module:}
The Grad-C module---Taylor binary cross entropy (TBCE) loss ($\mathcal{L}_{tbce}$)---is proposed to mitigate the larger gradient weights problem, which contracts the gradient weights of all wild data by Taylor series expansion in $\mathcal{L}_{bce}$. Simultaneously, the effectiveness of the $\mathcal{L}_{tbce}$ for rejecting unknown classes can be theoretically proven.

Given a function $t(x)$, if this function is differentiable to order $o$ at $x=x_0$, this function can be expressed as:
\begin{equation}\nonumber
    t(x)=\sum_{o=0}^{\infty}\frac{t^{o}(x_0)}{o!}(x-x_0)^o,
\end{equation}
where $t^{o}(x_0)$ is the $o$-th order derivative of $t(x)$ at $x=x_0$.

Let $t(f(\boldsymbol{x}))=-\log(1-f(\boldsymbol{x}))$ and $f(\boldsymbol{x}_{0})=0$. For $\forall o \geq 1$, the $t(f(\boldsymbol{x}))$ can be expressed as:
\begin{equation}\nonumber
    t(f(\boldsymbol{x}))=\sum_{o=1}^{\infty}\frac{f(\boldsymbol{x})^o}{o}.
\end{equation}
Then, the $\mathcal{L}_{tbce}$ is formalized as:
\begin{equation}
    \mathcal{L}_{tbce}(f(\boldsymbol{x}),y_{u})=-y_{u}\log(f(\boldsymbol{x}))+(1-y_{u})\sum_{o=1}^{t}\frac{{f(\boldsymbol{x})}^{o}}{o},
    \label{eq:taylot_binary_cross_entropy}
\end{equation}
where $t \in \mathcal{N}^+$ is the truncation order of the Taylor series expansion.

The gradient of $\mathcal{L}_{tbce}$ with respect to $\boldsymbol{\theta}$ for a wild sample ($\boldsymbol{x}_{wild},0$) is:
\begin{equation}
    \frac{\partial \mathcal{L}_{tbce}(f(\boldsymbol{x}_{wild}),0)}{\partial \boldsymbol{\theta}}={\frac{1-f(\boldsymbol{x}_{wild})^{t}}{1-f(\boldsymbol{x}_{wild})}}\nabla_{\boldsymbol{\theta}}f(\boldsymbol{x}_{wild}).
    \label{eq:gradient_wild_tbce}
\end{equation}
From Eqn.\ref{eq:gradient_wild_bce} and Eqn.\ref{eq:gradient_wild_tbce}, the gradient weight of a wild data in $\mathcal{L}_{tbce}$ is lower than that in $\mathcal{L}_{bcc}$, effectively mitigating the problem of larger gradient weights:
\begin{equation}
    {\frac{1-f(\boldsymbol{x}_{wild})^{t}}{1-f(\boldsymbol{x}_{wild})}} < {\frac{1}{1-f(\boldsymbol{x}_{wild})}}.
\end{equation}

Let $\mathcal{R}_{pu}(f)$ denote the risk where the unknown data is replaced by the wild data in $\mathcal{R}_{u}(f)$:
\begin{equation}
    \mathcal{R}_{pu}(f)=\frac{1}{2}\left(\mathcal{R}^{+}_{k}(f)+\mathcal{R}^{-}_{wild}(f)\right),
    \label{eq:unknown_classes_risk_pu}
\end{equation}
where $\mathcal{R}^{-}_{wild}(f)=\mathbb{E}_{(\boldsymbol{x}_{wild},0){\sim}{\mathbb{P}_{wild}}}\left[\mathcal{L}_{u}(f(\boldsymbol{x}_{wild}),0)\right]$. Let $\hat{f}$ and $f^{*}$ be the global minimizers of $\mathcal{R}_{pu}(f)$ and $\mathcal{R}_{u}(f)$, respectively. The theoretical property of $\mathcal{L}_{tbce}$ is stated in Theorem~\ref{theorem}, which demonstrates the reliability of estimating the rejection risk using wild data and $\mathcal{L}_{tbce}$. A detailed proof is provided in Appendix.
\begin{theorem}\label{theorem}
    The $\mathcal{R}_{u}(\hat{f})-\mathcal{R}_{u}(f^{*})$ and $\mathcal{R}_{pu}(f^{*})-\mathcal{R}_{pu}(\hat{f})$ are bounded when $\mathcal{L}_{tbce}$ is used as the loss function:
    \begin{equation}
        \begin{aligned}
            0 \leq {\mathcal{R}_{u}(\hat{f})-\mathcal{R}_{u}(f^*)} \leq {\pi\mathcal{N}_t},
        \end{aligned}
    \end{equation}
    \begin{equation}
        \begin{aligned}
            0 \leq {\mathcal{R}_{pu}(f^*)-\mathcal{R}_{pu}(\hat{f})} \leq {\pi\mathcal{N}_t},
        \end{aligned}
    \end{equation}
    where $\mathcal{N}_{t}={\sum_{o=1}^{t}}\frac{1}{o}$ is a constant releated to the truncation order $t$.
\end{theorem}

Based on this theoretical analysis, the performance of $\mathcal{R}_{pu}$ can be further improved by reducing both $\mathcal{N}_{t}$ and ${\pi}$. A lower $\mathcal{N}_{t}$ can be obtained by decreasing the truncation order $t$. Although the $\pi$ is a fixed constant for a given open-set HSI classification task, the original task can be decoupled into multiple sub-PU learning tasks via the proposed multi-PU head, and each sub-task would have a lower class prior.

\noindent \textbf{Grad-E Module:}
As previously mentioned, the Grad-C module reduces the gradient weights associated with both wild known and unknown samples, which would lead to $f$ underfitting the wild unknown data. To address this issue, as illustrated in Fig.~\ref{fig:grad}, the Grad-E module is designed to restore the gradient weights for the wild unknown data by weighting mechanism.

As illustrated in Fig.~\ref{fig:hopencls_framework}, two deep neural networks are used to contract and expand the gradient weights of wild data, respectively. In the Grad-E module, the $\mathcal{L}_{bce}$ is utilized to restore the gradient weights of wild unknown data with the confidence scores output by the network trained with Grad-C module. Moreover, the performance can be further enhanced by incorporating the confidence scores produced by the network trained with Grad-E module into the $\mathcal{L}_{tbce}$.

The weighted binary cross entropy (\textit{w}BCE) loss ($\mathcal{L}_{bce}^{w}$) is defined as follows:
\begin{equation}\nonumber
    \mathcal{L}^{w}_{bce}(f(\boldsymbol{x}),y_{u})=-y_{u}\log(f(\boldsymbol{x}))-(1-y_{u}){w_{e}}\log(1-f(\boldsymbol{x})),
    \label{eq:weight_binary_cross_entropy}
\end{equation}
where $w_{e}$ is the confidence score of data $\boldsymbol{x}$ output by the network trained with Grad-C module.
Similarly, the weighted Taylor binary cross entropy (\textit{w}TBCE) loss ($\mathcal{L}^{w}_{tbce}$) can be formulated as:
\begin{equation}\nonumber
    \mathcal{L}^{w}_{tbce}(f(\boldsymbol{x}),y_{u})=-y_{u}\log(f(\boldsymbol{x}))+(1-y_{u}){w_{c}}\sum_{o=1}^{t}\frac{{f(\boldsymbol{x})}^{o}}{o},
    \label{eq:weight_taylor_binary_cross_entropy}
\end{equation}
where $w_{c}$ is the confidence of data $\boldsymbol{x}$ output by the network trained with Grad-E module. The wild unknown samples are expected to receive higher confidence scores, with a maximum value of $1$. The following describes the way for obtaining the confidence scores, which involves both confidence score updates and probability mixing.

In the process of updating confidence scores, both $w_{c}$ and $w_{e}$ are optimized during model training, with an exponential moving average applied to stabilize the training. Two strategies are proposed for updating the confidence scores: \textbf{continuous updating} (Eqn.~\ref{eq:continuous_updating}) and \textbf{discrete updating} (Eqn.~\ref{eq:discrete_updating}):
\begin{equation}
    \begin{aligned}
    w_{c}={\alpha}w_{c}+(1-{\alpha})p_{e}\\
    w_{e}={\alpha}w_{e}+(1-{\alpha})p_{c},
    \end{aligned}
    \label{eq:continuous_updating}
\end{equation}
\begin{equation}
    \begin{aligned}
        &w_{c}={\alpha}w_{c}+(1-{\alpha})\mathbb{I}(p_{e} \geq \tau)\\
        &w_{e}={\alpha}w_{e}+(1-{\alpha})\mathbb{I}(p_{c} \geq \tau),
    \end{aligned}
    \label{eq:discrete_updating}
\end{equation}
where $\mathbb{I}(\bullet)$ is the indicator function, the initial values of $w_{c}$ and $w_{e}$ are both $1$, and $\tau$ is set to $0.95$. More experiments about $\tau$ are discussed in the experiment section. Experiments demonstrate that $w_{e}$ is more suitable for discrete updating and $w_{c}$ is better suited for continuous updating.

A straightforward way to compute $p_{c}$ and $p_{e}$ is by directly using the probability outputs from $f$ (\textbf{Pro}):
\begin{equation}
    \begin{aligned}
        p_{c}=1-f_{c}(\boldsymbol{x})\\
        p_{e}=1-f_{e}(\boldsymbol{x}),
    \end{aligned}
    \label{eq:pro}
\end{equation}
where $f_{c}(\boldsymbol{x})$ and $f_{e}(\boldsymbol{x})$ are the binary classifiers trained with $\mathcal{L}_{tbce}^{w}$ and $\mathcal{L}_{bce}^{w}$, respectively. However, this simplistic approach ignores the consistency between $q$ and $f$ (with multi-PU head): data from known classes should exhibit high probability outputs in both classifiers simultaneously. Therefore, a probability mixture strategy is proposed to incorporate the ability of classifying known classes of multi-class classifier into the multiple unknown rejection classifiers (\textbf{MixPro}):
\begin{equation}
    \begin{aligned}
        &p_{c}=1-(q_{c}^{c}(\boldsymbol{x}){\times}f_{c}^{c}(\boldsymbol{x}))\\
        &p_{e}=1-(q_{e}^{c}(\boldsymbol{x}){\times}f_{e}^{c}(\boldsymbol{x})),
    \end{aligned}
    \label{eq:mixpro}
\end{equation}
where $q_{\bullet}^{c}(\boldsymbol{x})$ is the probability output of the $c$-th class from the known classes classifier, $f_{\bullet}^{c}(\boldsymbol{x})$ is the probability output of the $c$-th sub-PU task. $q_{c}$ and $f_{c}$ share the same feature extractor, while $q_{e}$ and $f_{e}$ also share the same feature extractor. Moreover, the outputs of the $f_{c}^{c}(\boldsymbol{x})$ and $f_{e}^{c}(\boldsymbol{x})$ are aligned using KL divergence in the unknown rejection task, enabling the two networks to act as mutual teachers~\cite{T-HOneCls}.

\subsection{Overall Risk}

The risks for the two networks are defined as follows:
\begin{equation}
    \begin{aligned}
        &\hat{\mathcal{R}}_{c}(q_{c},f_{c})=\hat{\mathcal{R}}_{k}(q_{c})+\hat{\mathcal{R}}_{mpu}(f_{c},\mathcal{L}^{w}_{tbce})\\
        &\hat{\mathcal{R}}_{e}(q_{e},f_{e})=\hat{\mathcal{R}}_{k}(q_{e})+\hat{\mathcal{R}}_{mpu}(f_{e},\mathcal{L}^{w}_{bce}),
    \end{aligned}
\end{equation}
where $\hat{\mathcal{R}}_{mpu}(\bullet,\mathcal{L}^{w}_{tbce})$ and $\hat{\mathcal{R}}_{mpu}(\bullet,\mathcal{L}^{w}_{bce})$ represent that the loss function $\mathcal{L}^{w}_{tbce}$ and $\mathcal{L}^{w}_{bce}$ are used in Eqn.~\ref{eq:unknown_classes_risk_multi_pu_head_average_loss}, respectively.

The overall risk for the \textit{HOpnCls} framework can be formulated as follows:
\begin{equation}\nonumber
    \hat{\mathcal{R}}_{all}(q_{c},f_{c},q_{e},f_{e}) = \hat{\mathcal{R}}_{c}(q_{c},f_{c})+\hat{\mathcal{R}}_{e}(q_{e},f_{e})+\beta\hat{\mathcal{R}}_{kl}(f_{c},f_{e}),
\end{equation}
where the risk from KL divergence is denoted as $\hat{\mathcal{R}}_{kl}(f_{c},f_{e})$ and $\beta$ is a hyperparameter.

\section{Experimental Results}
\subsection{Experimental Settings}
\begin{figure*}[!t]
    \centering
    \includegraphics[width=0.98\textwidth]{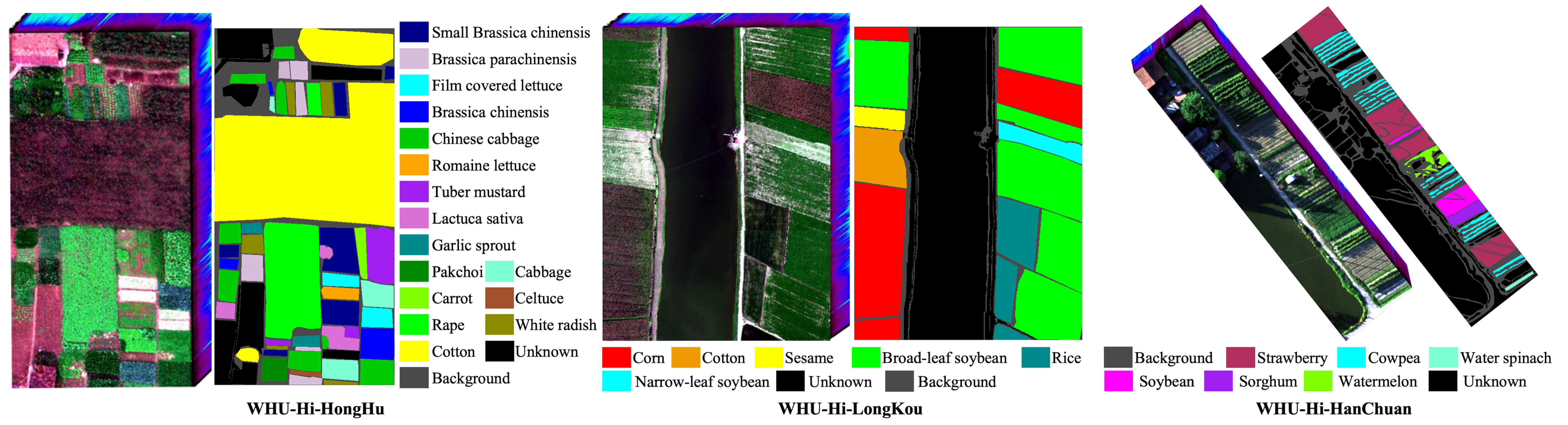}
    \caption{The WHU-Hi series HSI datasets: WHU-Hi-HongHu, WHU-Hi-LongKou, and WHU-Hi-HanChuan.}
    \label{fig:datasets}
\end{figure*}

\noindent \textbf{Datasets:}
To evaluate the performance of different methods, four HSI datasets are utilized: the University of Pavia dataset (Fig.~\ref{fig:open_set_example}) and the WHU-Hi series datasets~\cite{WHU-Hi} (Fig.~\ref{fig:datasets}): WHU-Hi-HongHu, WHU-Hi-LongKou, and WHU-Hi-Hanchuan. Notably, the WHU-Hi series datasets present significant spectral overlap~\cite{WHU-Hi}, making open-set HSI classification particularly challenging.

In the University of Pavia dataset, the original 9 classes are treated as known classes, while pixels that differ significantly from these known classes, such as buildings with notable reflective differences, are labeled as unknown classes, following the settings in~\cite{MDL4OW}. For the WHU-Hi series datasets, certain crop categories are considered as known classes, while other categories are labeled as unknown. This aligns with practical crop mapping scenarios, where the labeling workload is substantially reduced by focusing on crop categories, minimizing the need for annotating non-crop classes.

To simulate limited training sample conditions, only 100 samples per class are selected from each dataset for model training, with an additional 4000 wild samples randomly drawn from the imagery. Detailed information about the datasets is provided in Table~\ref{tab:datasets}.

\begin{table*}[!b]
    \centering
    \caption{Detailed information about datasets.}
    \label{tab:datasets}
    \resizebox{\linewidth}{!}{%
    \begin{tabular}{ccc|ccc|ccc|ccc|ccc} 
    \hline
    \multicolumn{6}{c|}{WHU-Hi-HongHu}                                                                                                                                                                                                                       & \multicolumn{3}{c|}{WHU-Hi-LongKou}                                              & \multicolumn{3}{c|}{WHU-Hi-HanChuan}                                       & \multicolumn{3}{c}{University of Pavia}                                   \\
    \rowcolor[rgb]{0.753,0.753,0.753} Index & Class Name                                                        & \# Sample & Index                                              & Class Name                                                    & \# Sample & Index & Class Name                                                   & \# Sample & Index & Class Name                                             & \# Sample & Index & Class Name                                           & \# Sample  \\ 
    \hline
    4                                       & Cotton                                                            & 164285    & 14                                                 & Lactuca sativa                                                & 7356      & 1     & Corn                                                         & 34511     & 1     & Strawberry                                             & 44735     & 1     & Asphalt                                              & 6631       \\
    6                                       & Rape                                                              & 44557     & 15                                                 & Celtuce                                                       & 1002      & 2     & Cotton                                                       & 8374      & 2     & Cowpea                                                 & 22753     & 2     & Meadows                                              & 18649      \\
    7                                       & \begin{tabular}[c]{@{}c@{}}Chinese\\cabbage\end{tabular}          & 24103     & 16                                                 & \begin{tabular}[c]{@{}c@{}}Film covered\\lettuce\end{tabular} & 7262      & 3     & Sesame                                                       & 3031      & 3     & Soybean                                                & 10287     & 3     & Gravel                                               & 2099       \\
    8                                       & Pakchoi                                                           & 4054      & 17                                                 & Romaine lettuce                                               & 3010      & 4     & \begin{tabular}[c]{@{}c@{}}Broad-leaf\\soybean\end{tabular}  & 63212     & 4     & Sorghum                                                & 5353      & 4     & Trees                                                & 3064       \\
    9                                       & Cabbage                                                           & 10819     & 18                                                 & Carrot                                                        & 3217      & 5     & \begin{tabular}[c]{@{}c@{}}Narrow-leaf\\soybean\end{tabular} & 4151      & 5     & \begin{tabular}[c]{@{}c@{}}Water\\spinach\end{tabular} & 1200      & 5     & \begin{tabular}[c]{@{}c@{}}Metal\\sheet\end{tabular} & 1345       \\
    10                                      & \begin{tabular}[c]{@{}c@{}}Tuber\\mustard\end{tabular}            & 12394     & 19                                                 & White radish                                                  & 8712      & 6     & Rice                                                         & 11854     & 6     & Watermelon                                             & 4533      & 6     & Bare soil                                            & 5029       \\
    11                                      & \begin{tabular}[c]{@{}c@{}}Brassica\\parachinensis\end{tabular}   & 11015     & 20                                                 & Garlic sprout                                                 & 3486      & 7-9   & Unknown                                                      & 79409     & 7-16  & Unknown                                                & 168669    & 7     & Bitumen                                              & 1330       \\
    12                                      & \begin{tabular}[c]{@{}c@{}}Brassica\\chinensis\end{tabular}       & 8954      & 21                                                 & Broad bean                                                    & 1328      &       &                                                              &           &       &                                                        &           & 8     & Brick                                                & 3682       \\
    13                                      & \begin{tabular}[c]{@{}c@{}}Small Brassica\\chinensis\end{tabular} & 22507     & \begin{tabular}[c]{@{}c@{}}1-3,\\5,22\end{tabular} & Unknown                                                       & 49632     &       &                                                              &           &       &                                                        &           & 9     & Shadow                                               & 947        \\
                                            &                                                                   &           &                                                    &                                                               &           &       &                                                              &           &       &                                                        &           & -     & Unknown                                              & 5163       \\
    \hline
    \end{tabular}}
\end{table*}

\noindent \textbf{Compared Methods:}
Two groups of methods are compared for thorough comparison: (1) trained with only $\mathbb{P}_{k}$ data: OpenMax~\cite{OpenMax}, CAC Loss~\cite{CACLoss}, MDL4OW~\cite{MDL4OW}, KL Matching~\cite{KLMatching}, MSP~\cite{MSP}, Energy~\cite{Energy} and Entropy~\cite{Entropy}; (2) trained with both $\mathbb{P}_{k}$ data and an additional dataset: DS3L~\cite{DS3L}, WOODS~\cite{WOODS}, OE~\cite{OE} and EOS~\cite{EOS}. Specifically, in addition to wild data, pure unknown classes data from $\mathbb{P}_{u}$ and wild data excluding unknown classes from $\mathbb{P}_{wild}-\mathbb{P}_{u}$ are used as the auxiliary data to demonstrate the robustness of \textit{HOpenCls} with respect to varying auxiliary data sources. To ensure a fair comparison, an equal amount of data (4000) from both distributions in the auxiliary dataset is randomly selected for model training.

Moreover, \textit{HOpenCls} is also compared with recently proposed PU learning methods (HOneCls~\cite{HOneCls} and T-HOneCls~\cite{T-HOneCls}) to demonstrate its effectiveness in PU learning tasks.

\noindent \textbf{Training Details:}
For patch-free methods, FreeNet~\cite{FPGA} is used as the spectral-spatial feature extractor. In the patch-based approaches, the encoder of FreeNet is used for the same purpose, with patch sizes increased to 9 to counteract the underutilization of spatial information caused by smaller spatial blocks. The \textit{HOpenCls} model was trained by 130 epoches (lr=$3{\times}10^{-4}$, momentum=0.9, weight\_decay=$10^{-4}$) with a ``CosineAnnealingLR'' learning rate policy. The default values for the weight $\beta$ and the Taylor expansion order $o$ were set to $1$ and $2$, respectively.

\noindent \textbf{Metrics:}
Following standard practice, Overall Accuracy (OA) is used to evaluate classifier performance. OA with and without unknown classes is referred to as Open OA and Closed OA, respectively. To evaluate the model’s ability to reject unknown classes, the F1 score (F1\textsuperscript{u}) and the Area Under the Curve (AUC\textsuperscript{u}) are also reported. All experiments were repeated 5 times, and the mean and standard deviation values are reported.

\subsection{Main Results}

\begin{figure*}[!t]
    \centering
    \includegraphics[width=0.98\textwidth]{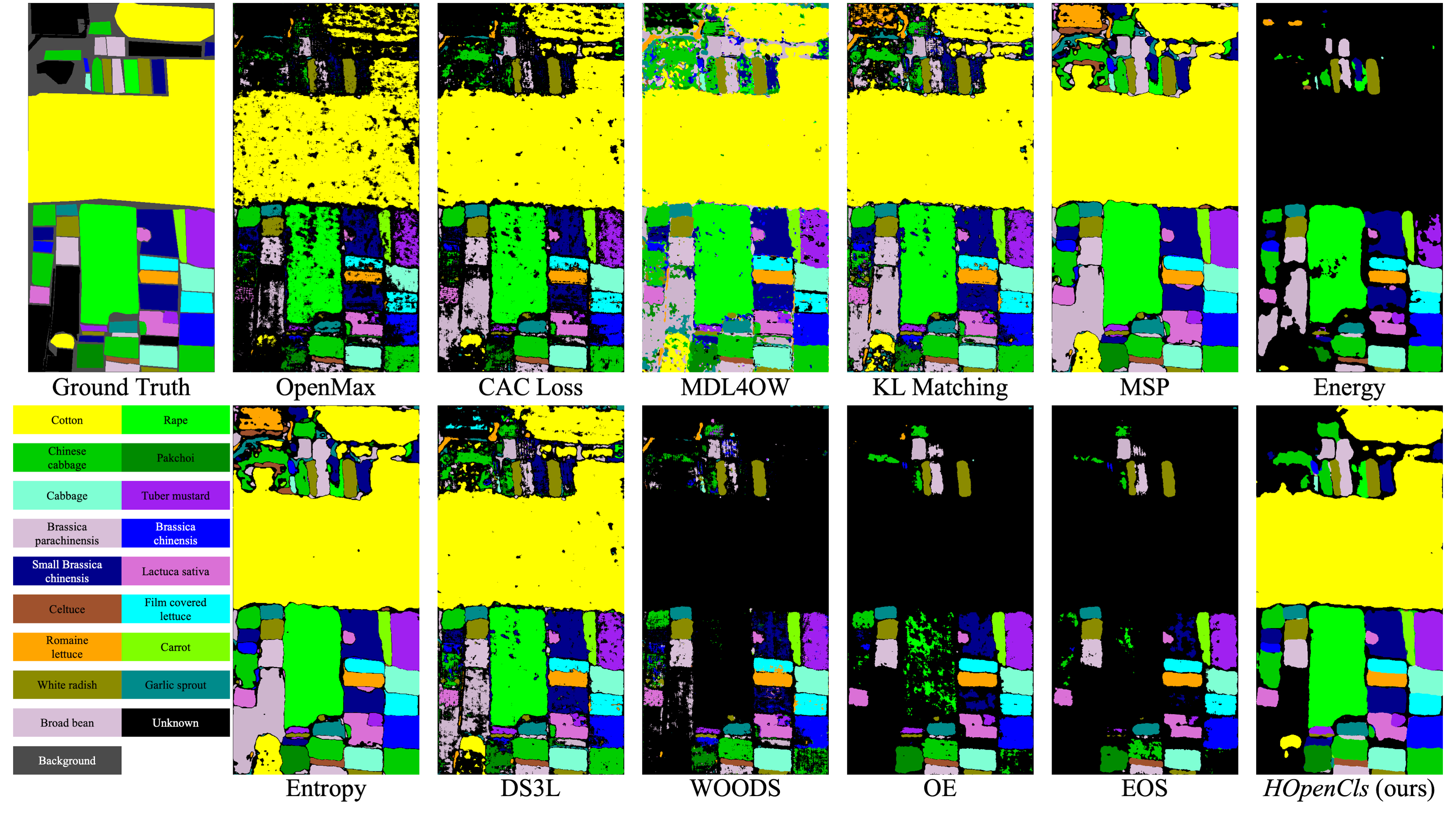}
    \caption{Open-set classification maps of WHU-Hi-HongHu dataset.}
    \label{fig:result_HH}
\end{figure*}

\begin{table*}[!t]
    \caption{Main results. ↑ indicates larger values are better. ±$x$ denotes the standard error.}
    \label{tab:main_results}
    \resizebox{\linewidth}{!}{%
    \begin{tabular}{lccccccccccccc} 
    \hline
    \multicolumn{1}{c}{\multirow{2}{*}{Methods}} & \multirow{2}{*}{$P$} & \multicolumn{3}{c}{WHU-Hi-HongHu}                                                                                  & \multicolumn{3}{c}{WHU-Hi-LongKou}                                                                                 & \multicolumn{3}{c}{WHU-Hi-HanChuan}                                                                                & \multicolumn{3}{c}{University of Pavia}                                                                             \\
    \multicolumn{1}{c}{}                         &                      & Open OA↑                             & F1\textsuperscript{u}↑               & AUC\textsuperscript{u}↑              & Open OA↑                             & F1\textsuperscript{u}↑               & AUC\textsuperscript{u}↑              & Open OA↑                             & F1\textsuperscript{u}↑               & AUC\textsuperscript{u}↑              & Open OA↑                             & F1\textsuperscript{u}↑               & AUC\textsuperscript{u}↑               \\ 
    \hline
                                                 & \multicolumn{1}{l}{} & \multicolumn{12}{c}{{\cellcolor[rgb]{0.753,0.753,0.753}}With known classes data from~$\mathbb{P}_{k}$ only}                                                                                                                                                                                                                                                                                                                                                                                               \\
    OpenMax~\cite{OpenMax}                                      & $P_{pb}$               & 73.84\textsuperscript{±2.0}          & 46.88\textsuperscript{±2.0}          & 86.63\textsuperscript{±2.8}          & 88.03\textsuperscript{±0.6}          & 86.54\textsuperscript{±0.6}          & 98.30\textsuperscript{±0.3}          & 74.82\textsuperscript{±2.7}          & 80.00\textsuperscript{±2.6}          & 83.02\textsuperscript{±2.1}          & 74.41\textsuperscript{±1.0}          & 44.00\textsuperscript{±3.3}          & 81.33\textsuperscript{±3.4}           \\
    CAC Loss~\cite{CACLoss}                                     & $P_{pb}$               & 79.16\textsuperscript{±0.6}          & 46.71\textsuperscript{±2.0}          & 84.04\textsuperscript{±2.5}          & 58.72\textsuperscript{±1.0}          & 20.45\textsuperscript{±0.3}          & 15.33\textsuperscript{±1.0}          & 62.24\textsuperscript{±7.2}          & 67.35\textsuperscript{±10.8}         & 69.94\textsuperscript{±6.8}          & 86.91\textsuperscript{±1.8}          & 55.85\textsuperscript{±7.4}          & 92.38\textsuperscript{±2.7}           \\
    MDL4OW~\cite{MDL4OW}                                       & $P_{pb}$               & 84.14\textsuperscript{±0.3}          & 20.08\textsuperscript{±3.7}          & 56.43\textsuperscript{±0.7}          & 95.69\textsuperscript{±0.1}          & 95.37\textsuperscript{±0.2}          & 97.80\textsuperscript{±0.1}          & 36.42\textsuperscript{±0.2}          & 12.51\textsuperscript{±0.5}          & 48.20\textsuperscript{±0.2}          & 88.94\textsuperscript{±0.3}          & 25.32\textsuperscript{±3.6}          & 61.37\textsuperscript{±0.5}           \\
    KL Matching~\cite{KLMatching}                                  & $P_{pb}$               & 83.07\textsuperscript{±0.7}          & 39.62\textsuperscript{±5.6}          & 77.40\textsuperscript{±5.1}          & 59.94\textsuperscript{±0.7}          & 11.74\textsuperscript{±1.6}          & 76.93\textsuperscript{±2.1}          & 53.24\textsuperscript{±7.9}          & 48.24\textsuperscript{±14.7}         & 77.64\textsuperscript{±5.2}          & 89.44\textsuperscript{±1.6}          & 52.10\textsuperscript{±11.0}         & 86.96\textsuperscript{±8.4}           \\
    MSP~\cite{MSP}                                          & $P_{pf}$               & 87.44\textsuperscript{±0.7}          & 31.00\textsuperscript{±6.2}          & 81.01\textsuperscript{±10.1}         & 87.74\textsuperscript{±5.1}          & 81.35\textsuperscript{±9.3}          & 98.02\textsuperscript{±0.6}          & 39.76\textsuperscript{±0.9}          & 17.11\textsuperscript{±2.3}          & 72.69\textsuperscript{±3.1}          & 93.07\textsuperscript{±1.0}          & 58.37\textsuperscript{±8.5}          & 97.96\textsuperscript{±0.6}           \\
    Energy~\cite{Energy}                                       & $P_{pf}$               & 37.94\textsuperscript{±3.5}          & 25.61\textsuperscript{±2.7}          & 70.21\textsuperscript{±4.9}          & 88.77\textsuperscript{±2.4}          & 87.11\textsuperscript{±2.4}          & 97.45\textsuperscript{±0.9}          & 72.14\textsuperscript{±1.5}          & 80.22\textsuperscript{±1.1}          & 75.28\textsuperscript{±1.9}          & 83.60\textsuperscript{±4.4}          & 56.81\textsuperscript{±7.1}          & 96.01\textsuperscript{±2.5}           \\
    Entropy~\cite{Entropy}                                      & $P_{pf}$               & 87.83\textsuperscript{±0.8}          & 37.20\textsuperscript{±6.5}          & 80.88\textsuperscript{±10.0}         & 92.44\textsuperscript{±4.2}          & 89.40\textsuperscript{±6.7}          & 98.24\textsuperscript{±0.6}          & 42.24\textsuperscript{±1.3}          & 23.65\textsuperscript{±3.1}          & 73.19\textsuperscript{±3.2}          & 94.12\textsuperscript{±1.0}          & 67.83\textsuperscript{±7.0}          & 97.99\textsuperscript{±0.6}           \\ 
    \hline
                                                 &                      & \multicolumn{12}{c}{{\cellcolor[rgb]{0.753,0.753,0.753}}With known classes data from~$\mathbb{P}_{k}$ and wild data from~$\mathbb{P}_{wild}$}                                                                                                                                                                                                                                                                                                                                                                            \\
    DS3L~\cite{DS3L}                                         & $P_{pb}$               & 83.93\textsuperscript{±0.7}          & 48.70\textsuperscript{±2.9}          & 80.06\textsuperscript{±4.4}          & 93.86\textsuperscript{±0.2}          & 92.99\textsuperscript{±0.5}          & 98.65\textsuperscript{±0.3}          & 72.37\textsuperscript{±1.0}          & 75.25\textsuperscript{±1.3}          & 90.80\textsuperscript{±1.0}          & 91.21\textsuperscript{±1.9}          & 60.22\textsuperscript{±14.7}         & 92.24\textsuperscript{±8.6}           \\
    WOODS~\cite{WOODS}                                        & $P_{pb}$               & 43.03\textsuperscript{±18.9}         & 32.67\textsuperscript{±11.2}         & 60.48\textsuperscript{±20.0}         & 76.67\textsuperscript{±15.7}         & 78.44\textsuperscript{±12.7}         & 98.26\textsuperscript{±1.3}          & 86.33\textsuperscript{±4.3}          & 90.69\textsuperscript{±2.7}          & 94.89\textsuperscript{±2.3}          & 59.87\textsuperscript{±24.5}         & 41.49\textsuperscript{±21.2}         & 91.94\textsuperscript{±6.9}           \\
    OE~\cite{OE}                                           & $P_{pf}$               & 36.36\textsuperscript{±0.6}          & 28.62\textsuperscript{±0.2}          & 93.21\textsuperscript{±0.9}          & 78.25\textsuperscript{±1.4}          & 78.15\textsuperscript{±1.1}          & 99.12\textsuperscript{±0.1}          & 92.55\textsuperscript{±0.6}          & 94.58\textsuperscript{±0.4}          & 98.84\textsuperscript{±0.2}          & 80.14\textsuperscript{±3.3}          & 52.41\textsuperscript{±3.9}          & 97.66\textsuperscript{±0.3}           \\
    EOS~\cite{EOS}                                          & $P_{pf}$               & 31.75\textsuperscript{±0.2}          & 27.26\textsuperscript{±0.1}          & 89.25\textsuperscript{±1.1}          & 52.13\textsuperscript{±0.5}          & 61.86\textsuperscript{±0.2}          & 99.10\textsuperscript{±0.1}          & 82.75\textsuperscript{±1.0}          & 88.35\textsuperscript{±0.6}          & 98.76\textsuperscript{±0.2}          & 58.37\textsuperscript{±2.5}          & 34.32\textsuperscript{±1.3}          & 96.93\textsuperscript{±0.7}           \\
    \textit{HOpenCls} (ours)                                     & $P_{pf}$               & \textbf{96.03}\textsuperscript{±0.4} & \textbf{87.61}\textsuperscript{±1.2} & \textbf{99.22}\textsuperscript{±0.2} & \textbf{99.15}\textsuperscript{±0.1} & \textbf{99.03}\textsuperscript{±0.1} & \textbf{99.99}\textsuperscript{±0.0} & \textbf{97.54}\textsuperscript{±0.3} & \textbf{98.14}\textsuperscript{±0.2} & \textbf{99.67}\textsuperscript{±0.1} & \textbf{97.17}\textsuperscript{±0.2} & \textbf{88.76}\textsuperscript{±0.7} & \textbf{99.75}\textsuperscript{±0.0}  \\
    \hline
    \end{tabular}
    }
\end{table*}

\begin{table*}[!t]
    \centering
    \caption{Closed OA for different methods. ±$x$ denotes the standard error.}
    \label{tab:closed_OA}
    \resizebox{\linewidth}{!}{%
    \begin{tabular}{l|ccccccc|ccccc} 
    \hline
    \multicolumn{1}{c|}{\multirow{3}{*}{Datasets}} & \multicolumn{7}{c|}{{\cellcolor[rgb]{0.753,0.753,0.753}}With known classes data from~$\mathbb{P}_{k}$only}                                                                                                                              & \multicolumn{5}{c}{{\cellcolor[rgb]{0.753,0.753,0.753}}With known classes data from~$\mathbb{P}_{k}$~and wild data from~$\mathbb{P}_{wild}$}                                                         \\
    \multicolumn{1}{c|}{}                          & OpenMax                     & CAC Loss                    & MDL4OW                      & KL Matching                 & MSP                         & Energy                      & Entropy                     & DS3L                        & WOODS                       & OE                          & EOS                         & \textit{HOpenCls} (ours)                              \\
    \multicolumn{1}{c|}{}                          & $P_{pb}$                    & $P_{pb}$                    & $P_{pb}$                    & $P_{pb}$                    & $P_{pf}$                    & $P_{pf}$                    & $P_{pf}$                    & $P_{pb}$                    & $P_{pb}$                    & $P_{pf}$                    & $P_{pf}$                    & $P_{pf}$                              \\ 
    \hline
    WHU-Hi-HongHu                                  & 93.78\textsuperscript{±0.6} & 94.64\textsuperscript{±0.1} & 95.93\textsuperscript{±0.1} & 93.78\textsuperscript{±0.6} & 98.23\textsuperscript{±0.1} & 98.23\textsuperscript{±0.1} & 98.23\textsuperscript{±0.1} & 93.93\textsuperscript{±0.2} & 90.98\textsuperscript{±1.6} & 96.68\textsuperscript{±0.2} & 96.79\textsuperscript{±0.3} & \textbf{98.57}\textsuperscript{±0.2}  \\
    WHU-Hi-LongKou                                 & 96.97\textsuperscript{±0.4} & 96.99\textsuperscript{±0.3} & 98.64\textsuperscript{±0.1} & 96.97\textsuperscript{±0.4} & 99.37\textsuperscript{±0.1} & 99.37\textsuperscript{±0.1} & 99.37\textsuperscript{±0.1} & 97.17\textsuperscript{±0.4} & 94.89\textsuperscript{±1.9} & 94.46\textsuperscript{±0.7} & 95.14\textsuperscript{±1.1} & \textbf{99.75}\textsuperscript{±0.0}  \\
    WHU-Hi-HanChuan                                & 94.37\textsuperscript{±0.4} & 95.42\textsuperscript{±0.5} & 96.74\textsuperscript{±0.3} & 94.37\textsuperscript{±0.4} & 98.50\textsuperscript{±0.2} & 98.50\textsuperscript{±0.2} & 98.50\textsuperscript{±0.2} & 95.03\textsuperscript{±0.4} & 93.59\textsuperscript{±2.0} & 98.28\textsuperscript{±0.2} & 98.17\textsuperscript{±0.7} & \textbf{99.77}\textsuperscript{±0.1}  \\
    University of Pavia                            & 96.70\textsuperscript{±0.6} & 98.21\textsuperscript{±0.2} & 98.72\textsuperscript{±0.3} & 96.70\textsuperscript{±0.6} & 99.65\textsuperscript{±0.1} & 99.65\textsuperscript{±0.1} & 99.65\textsuperscript{±0.1} & 97.85\textsuperscript{±0.2} & 96.31\textsuperscript{±1.4} & 99.25\textsuperscript{±0.2} & 99.27\textsuperscript{±0.1} & \textbf{99.71}\textsuperscript{±0.1}  \\
    \hline
    \end{tabular}
    }
\end{table*}

\begin{figure*}[!t]
    \centering
    \includegraphics[width=0.98\textwidth]{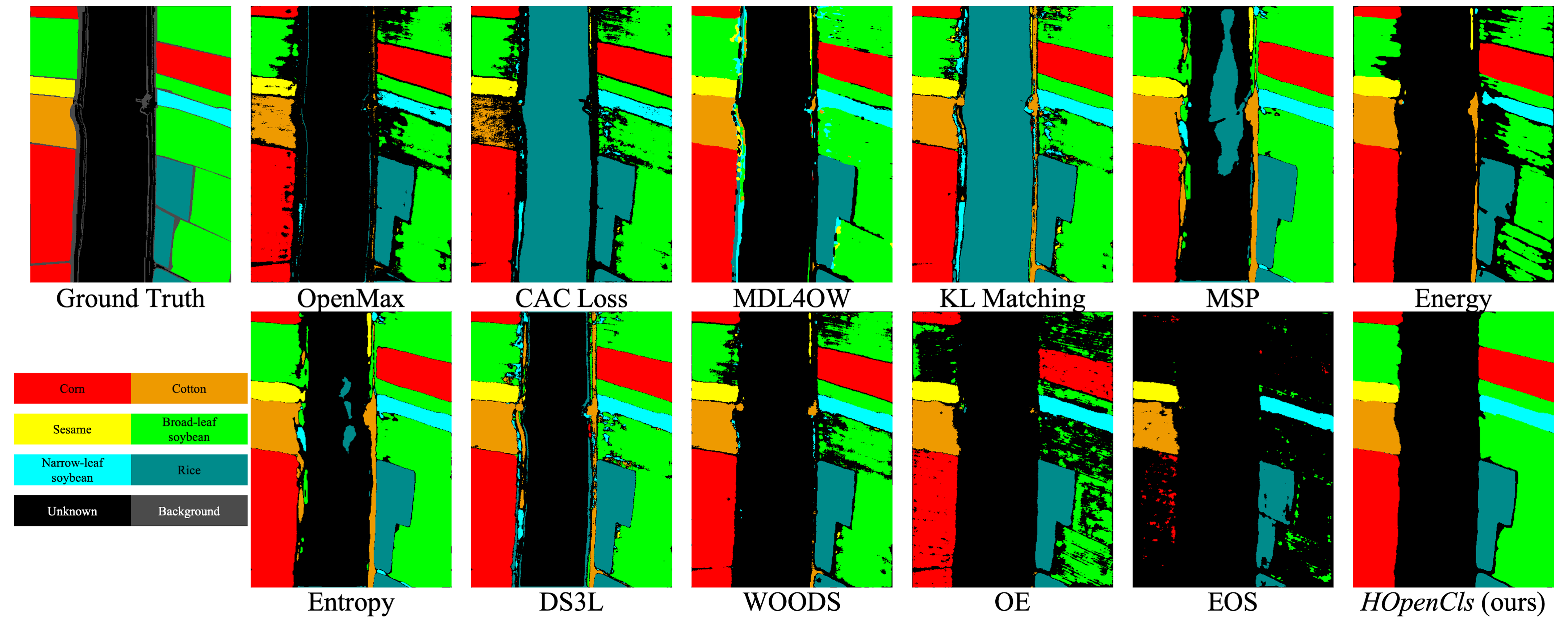}
    \caption{Open-set classification maps of WHU-Hi-LongKou dataset.}
    \label{fig:result_LK}
\end{figure*}

\begin{figure*}[!t]
    \centering
    \includegraphics[width=0.98\textwidth]{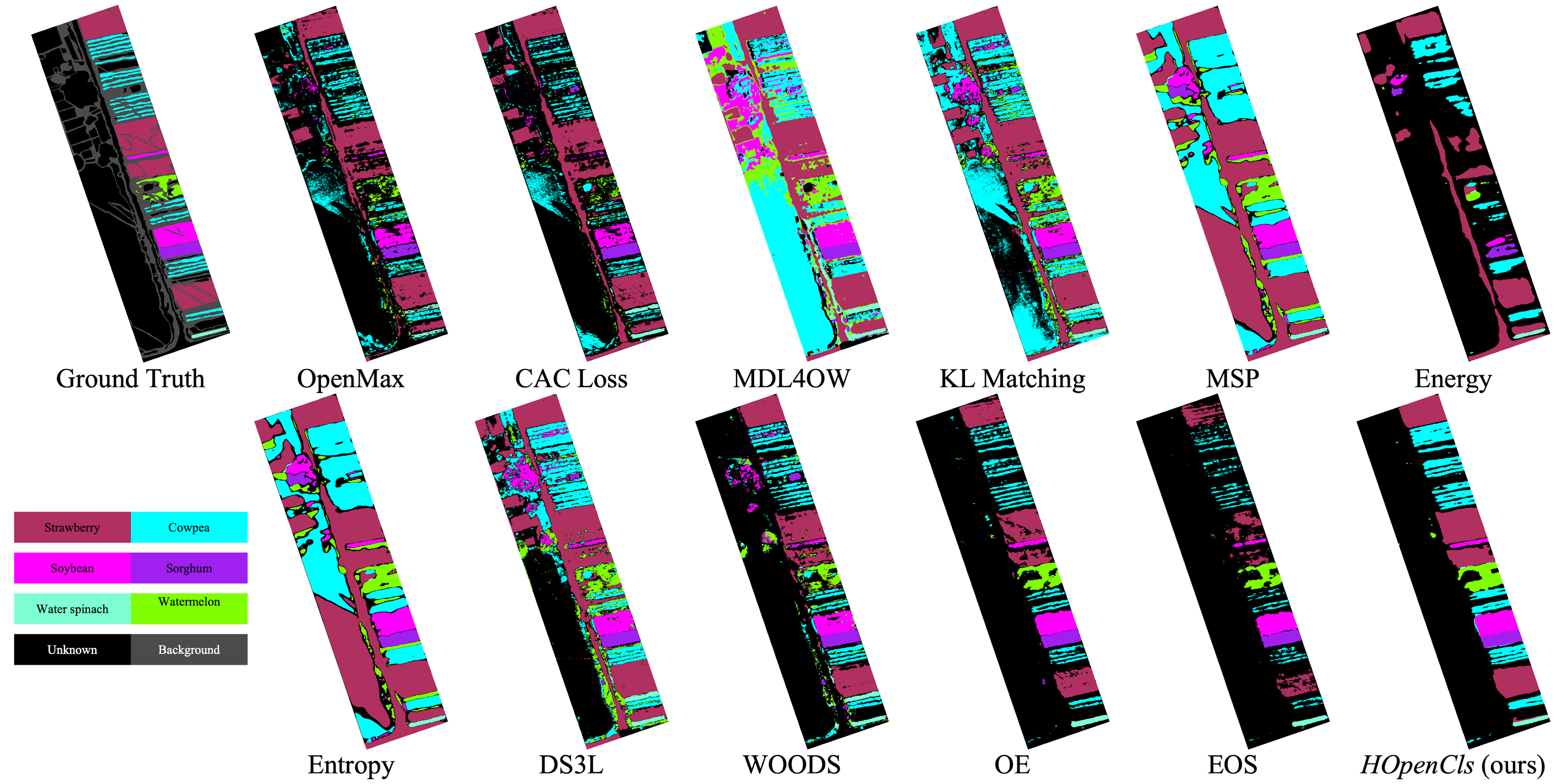}
    \caption{Open-set classification maps of WHU-Hi-HanChuan dataset.}
    \label{fig:result_HC}
\end{figure*}

\begin{figure*}[!t]
    \centering
    \includegraphics[width=0.98\textwidth]{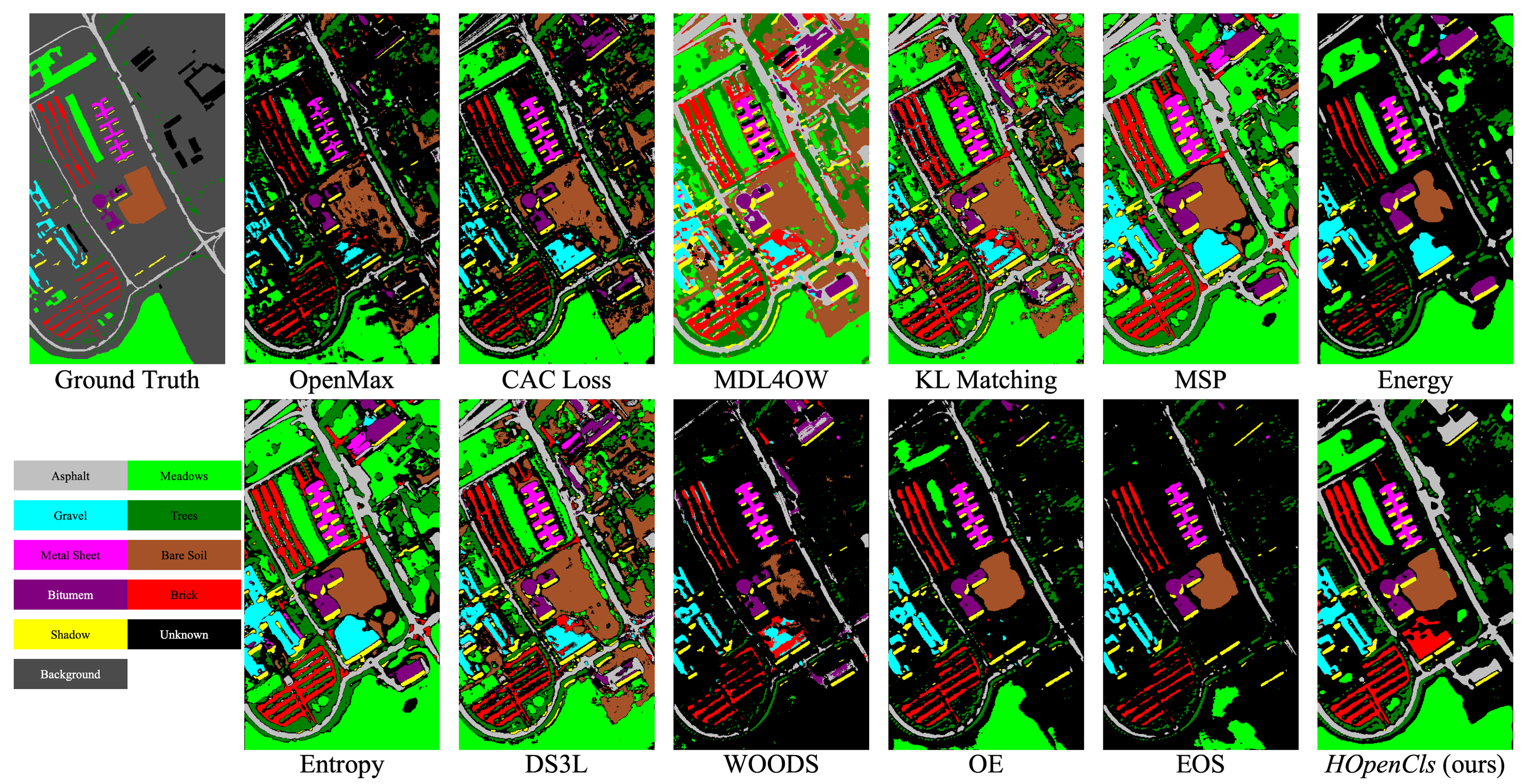}
    \caption{Open-set classification maps of University of Pavia dataset.}
    \label{fig:result_PU}
\end{figure*}

\noindent \textbf{\textit{HOpenCls} Achieves Superior Performance:}
\textit{HOpenCls} achieves the highest scores across all metrics and datasets. The results for Open OA, F1\textsuperscript{u}, and AUC\textsuperscript{u} are detailed in Table~\ref{tab:main_results}, while the closed OA results are presented in Table~\ref{tab:closed_OA}. Classification maps generated by different methods can be found in Fig.~\ref{fig:result_HH}-Fig.~\ref{fig:result_PU}. Compared to the second-best method, \textit{HOpenCls} shows improvements in Open OA of 8.20, 3.46, 4.99, and 3.05 on the WHU-Hi-HongHu, WHU-Hi-LongKou, WHU-Hi-HanChuan, and University of Pavia, respectively. Notably, for unknown class rejection tasks, \textit{HOpenCls} exhibits significant improvements in F1\textsuperscript{u}, with gains of 38.91, 3.66, 3.56, and 20.93 on the same datasets, respectively.

\begin{table*}[!t]
    \caption{Results of classifiers with extra data from different distributions. ↑ indicates larger values are better. ±$x$ denotes the standard error.}
    \label{tab:additional_data}
    \resizebox{\linewidth}{!}{%
    \begin{tabular}{lccccccccccccc} 
    \hline
    \multicolumn{1}{c}{\multirow{2}{*}{Methods}} & \multirow{2}{*}{$P$}         & \multicolumn{3}{c}{WHU-Hi-HongHu}                                                                                   & \multicolumn{3}{c}{WHU-Hi-LongKou}                                                                                  & \multicolumn{3}{c}{WHU-Hi-HanChuan}                                                                                & \multicolumn{3}{c}{University of Pavia}                                                                              \\
    \multicolumn{1}{c}{}                         &                              & Open OA↑                             & F1\textsuperscript{u}↑               & AUC\textsuperscript{u}↑               & Open OA↑                             & F1\textsuperscript{u}↑               & AUC\textsuperscript{u}↑               & Open OA↑                             & F1\textsuperscript{u}↑               & AUC\textsuperscript{u}↑              & Open OA↑                             & F1\textsuperscript{u}↑               & AUC\textsuperscript{u}↑                \\ 
    \hline
                                                 & \multicolumn{1}{l}{}         & \multicolumn{12}{c}{{\cellcolor[rgb]{0.753,0.753,0.753}}With known classes data from~$\mathbb{P}_{k}$ and pure unknown classes data from~$\mathbb{P}_{u}$}                                                                                                                                                                                                                                                                                                                                                                                   \\
    DS3L~\cite{DS3L}                                         & \multicolumn{1}{l}{$P_{pb}$} & 89.08\textsuperscript{±0.7}          & 71.64\textsuperscript{±2.1}          & 96.73\textsuperscript{±0.6}           & 94.56\textsuperscript{±0.5}          & 94.05\textsuperscript{±0.6}          & 99.25\textsuperscript{±0.1}           & 73.64\textsuperscript{±2.1}          & 76.79\textsuperscript{±2.4}          & 93.32\textsuperscript{±0.6}          & 95.50\textsuperscript{±0.5}          & 84.92\textsuperscript{±1.4}          & 99.54\textsuperscript{±0.2}            \\
    WOODS~\cite{WOODS}                                        & $P_{pb}$                     & 74.48\textsuperscript{±11.4}         & 55.31\textsuperscript{±11.0}         & 99.28\textsuperscript{±0.4}           & 92.70\textsuperscript{±1.2}          & 91.51\textsuperscript{±1.3}          & 99.86\textsuperscript{±0.1}           & 86.71\textsuperscript{±4.8}          & 90.50\textsuperscript{±4.2}          & 95.00\textsuperscript{±3.4}          & 89.47\textsuperscript{±3.9}          & 74.72\textsuperscript{±9.0}          & 99.95\textsuperscript{±0.1}            \\
    OE~\cite{OE}                                           & $P_{pf}$                     & 95.23\textsuperscript{±0.6}          & 86.49\textsuperscript{±1.8}          & 99.99\textsuperscript{±0.0}           & 98.83\textsuperscript{±0.2}          & 98.66\textsuperscript{±0.3}          & 99.99\textsuperscript{±0.0}           & \textbf{98.85}\textsuperscript{±0.2} & \textbf{99.15}\textsuperscript{±0.1} & 99.94\textsuperscript{±0.0}          & 98.11\textsuperscript{±0.4}          & 92.59\textsuperscript{±1.4}          & \textbf{100.00}\textsuperscript{±0.0}  \\
    EOS~\cite{EOS}                                          & $P_{pf}$                     & 95.45\textsuperscript{±0.3}          & 88.28\textsuperscript{±1.2}          & 99.98\textsuperscript{±0.0}           & 98.44\textsuperscript{±0.2}          & 98.19\textsuperscript{±0.2}          & \textbf{100.00}\textsuperscript{±0.0} & 98.63\textsuperscript{±0.1}          & 98.99\textsuperscript{±0.1}          & \textbf{99.94}\textsuperscript{±0.0} & 97.49\textsuperscript{±0.7}          & 90.36\textsuperscript{±2.7}          & \textbf{100.00}\textsuperscript{±0.0}  \\
    \textit{HOpenCls} (ours)                                     & $P_{pf}$                     & \textbf{96.95}\textsuperscript{±0.2} & \textbf{98.98}\textsuperscript{±0.4} & \textbf{100.00}\textsuperscript{±0.0} & \textbf{99.10}\textsuperscript{±0.2} & \textbf{99.91}\textsuperscript{±0.0} & \textbf{100.00}\textsuperscript{±0.0} & 98.15\textsuperscript{±0.3}          & 98.73\textsuperscript{±0.2}          & 99.73\textsuperscript{±0.1}          & \textbf{99.34}\textsuperscript{±0.1} & \textbf{99.69}\textsuperscript{±0.0} & \textbf{100.00}\textsuperscript{±0.0}  \\ 
    \hline
                                                 &                              & \multicolumn{12}{c}{{\cellcolor[rgb]{0.753,0.753,0.753}}With known classes data from~$\mathbb{P}_{k}$ and wild data excluding unknown classes from~$\mathbb{P}_{wild}-\mathbb{P}_{u}$}                                                                                                                                                                                                                                                                                                                                                                 \\
    DS3L~\cite{DS3L}                                          & $P_{pb}$                     & 82.26\textsuperscript{±0.7}          & 41.83\textsuperscript{±3.0}          & 71.59\textsuperscript{±4.2}           & 60.37\textsuperscript{±0.6}          & 14.46\textsuperscript{±1.5}          & 15.24\textsuperscript{±1.7}           & 54.87\textsuperscript{±8.7}          & 50.76\textsuperscript{±14.0}         & 81.89\textsuperscript{±1.8}          & 91.21\textsuperscript{±1.3}          & 60.07\textsuperscript{±8.7}          & 93.54\textsuperscript{±4.0}            \\
    WOODS~\cite{WOODS}                                        & $P_{pb}$                     & 31.87\textsuperscript{±9.2}          & 16.42\textsuperscript{±3.6}          & 34.81\textsuperscript{±11.6}          & 53.61\textsuperscript{±23.6}         & 47.23\textsuperscript{±29.7}         & 52.47\textsuperscript{±26.8}          & 77.14\textsuperscript{±5.5}          & 84.32\textsuperscript{±3.3}          & 87.45\textsuperscript{±3.4}          & 44.12\textsuperscript{±6.5}          & 26.29\textsuperscript{±3.6}          & 75.66\textsuperscript{±7.5}            \\
    OE~\cite{OE}                                           & $P_{pf}$                     & 36.16\textsuperscript{±0.6}          & 27.29\textsuperscript{±0.7}          & 71.16\textsuperscript{±4.6}           & 65.40\textsuperscript{±3.2}          & 69.09\textsuperscript{±2.0}          & 93.36\textsuperscript{±2.1}           & 82.68\textsuperscript{±1.3}          & 87.80\textsuperscript{±0.9}          & 92.02\textsuperscript{±1.1}          & 78.86\textsuperscript{±2.8}          & 49.79\textsuperscript{±2.9}          & 95.53\textsuperscript{±0.5}            \\
    EOS~\cite{EOS}                                          & $P_{pf}$                     & 30.83\textsuperscript{±0.7}          & 25.53\textsuperscript{±1.2}          & 53.76\textsuperscript{±3.2}           & 49.96\textsuperscript{±0.4}          & 60.74\textsuperscript{±0.2}          & 94.06\textsuperscript{±1.8}           & 73.07\textsuperscript{±0.4}          & 82.82\textsuperscript{±0.3}          & 91.44\textsuperscript{±1.0}          & 54.38\textsuperscript{±2.6}          & 32.22\textsuperscript{±1.3}          & 95.22\textsuperscript{±1.3}            \\
    \textit{HOpenCls} (ours)                                     & $P_{pf}$                     & \textbf{92.77}\textsuperscript{±0.8} & \textbf{71.58}\textsuperscript{±4.2} & \textbf{95.30}\textsuperscript{±1.5}  & \textbf{98.77}\textsuperscript{±0.2} & \textbf{98.50}\textsuperscript{±0.2} & \textbf{99.16}\textsuperscript{±0.3}  & \textbf{87.37}\textsuperscript{±5.4} & \textbf{89.29}\textsuperscript{±5.2} & \textbf{97.91}\textsuperscript{±0.6} & \textbf{97.09}\textsuperscript{±0.4} & \textbf{88.48}\textsuperscript{±1.4} & \textbf{99.69}\textsuperscript{±0.1}   \\
    \hline
    \end{tabular}
    }
\end{table*}

Several observations should be highlighted: (1) A significant bottleneck in the development of HSI classifiers is the challenge of rejecting unknown classes. Different classifiers show substantial variation in their performance on this task, take the WHU-Hi-HongHu HSI dataset as an example, with some suffering from under-recognition (e.g., KL Matching, MSP, Entropy, and DS3L) and others from over-recognition (e.g., Energy, WOODS, OE, and EOS). The proposed \textit{HOpenCls} effectively balances these issues, achieving a 38.91 and 20.93 improvement in F1\textsuperscript{u} on the WHU-Hi-HongHu and University of Pavia datasets, respectively. (2) \textit{HOpenCls} not only excels at unknown classes rejection but also improves classification performance on known classes. The task of unknown classes rejection has the ability to improve the classification performance of known classes in the proposed framework, which exhibits the effectiveness of the multi-task architecture. (3) The use of additional data can negatively affect classifier performance. For example, on the WHU-Hi-HongHu dataset, WOODS underperforms compared to OpenMax and CAC Loss, both of which do not incorporate additional data. Additionally, classifiers like OE and EOS experience significant degradation due to the influence of known class components in wild data. This suggests that while wild data is easily accessible, it introduces greater challenges for algorithm design.

\noindent \textbf{\textit{HOpenCls} is Robust to the Sources of Auxiliary Dataset:}
Three types of extra datasets are used to evaluate the robustness of the proposed \textit{HOpenCls} concerning the sources of the extra dataset. Wild data, which is the focus of this paper, can be obtained from the living environments of the classifiers with almost no cost, the results are shown in Table~\ref{tab:main_results}. Pure unknown classes data represents the ideal extra data but is labor-intensive to obtain, originating from the distribution of $\mathbb{P}_{u}$. Wild data without unknown classes poses the greatest challenge, providing no information about unknown classes and coming from the distribution of $\mathbb{P}_{wild}-\mathbb{P}_{u}$. The results of the models trained with pure unknown data or wild data excluding unknown classes are shown in Table~\ref{tab:additional_data}.

Some conclusions can be drawn from the results: (1) \textit{HOpenCls} demonstrates robustness to the source of extra data, and the performance of the proposed model does not significantly decrease compared to other methods when wild data excluding unknown classes is used as extra data. (2) The use of pure unknown data significantly enhances performance, indicating that it is the most effective form of extra data. However, acquiring unknown data is labor-intensive, and the collected unknown data is usually difficult to cover all the unknown classes. (3) Wild data is the potential extra data that can be almost freely collected, but the wild data propose higher demands on the algorithm design due to the wild data is the mixture of known and unknown classes. Due to the influence of known classes in the wild data, both OE and EOS suffer from significant performance degradation. (4) The proposed \textit{HOpenCls} achieves the best metrics across all datasets when using wild data excluding unknown classes, suggesting that it excels at discovering novel classes because this dataset does not contain any information about unknown classes.

\noindent \textbf{The combination of Grad-C and Grad-E Modules is a Powerful Classifier for PU Learning:}
Compared to other HSI PU learning methods, the combination of Grad-C and Grad-E modules, referred to as \textit{HOpenCls(PU)}, achieves better performance in PU learning tasks (Table~\ref{tab:pu_results}). Due to the limitations of inaccurate class priors, the performance of HOneCls significantly decreases. Compared to the class prior-free method (T-HOneCls), the proposed method gets better performance, particularly in high class prior scenarios, such as for Cotton and Broad-leaf soybean.

\begin{table}[!t]
    \centering
    \caption{F1 scores for different PU learning methods. ±$x$ denotes the standard error.}
    \label{tab:pu_results}
    \resizebox{\linewidth}{!}{%
    \begin{tabular}{cccccc} 
    \hline
    \multirow{3}{*}{Methods\tablefootnote{The experimental settings are consistent with those outlined in~\cite{T-HOneCls}.}} & \multicolumn{5}{c}{Positive class (class prior)}                                                                                                                                                                                    \\
                                                                                                                              & Cotton\tablefootnote{The Cotton class from the WHU-Hi-HongHu dataset.} & Broad-leaf soybean                   & Corn                                 & Rape                                 & Cowpea                                \\
                                                                                                                              & (0.3769)                                                               & (0.2873)                             & (0.1569)                             & (0.1317)                             & (0.0617)                              \\ 
    \hline
    HOneCls~\cite{HOneCls}                                                                                                                   & \textbf{99.44}\textsuperscript{±0.2}                                   & 88.02\textsuperscript{±0.3}          & 99.67\textsuperscript{±0.1}          & 81.81\textsuperscript{±1.2}          & 58.97\textsuperscript{±3.6}           \\
    T-HOneCls~\cite{T-HOneCls}                                                                                                                 & 98.15\textsuperscript{±0.3}                                            & 92.64\textsuperscript{±0.9}          & 99.70\textsuperscript{±0.1}          & 97.81\textsuperscript{±0.2}          & 90.31\textsuperscript{±1.1}           \\
    \textit{HOpenCls(PU)}                                                                                                              & 99.33\textsuperscript{±0.2}                                            & \textbf{94.98}\textsuperscript{±1.2} & \textbf{99.72}\textsuperscript{±0.0} & \textbf{98.40}\textsuperscript{±0.2} & \textbf{90.67}\textsuperscript{±2.1}  \\
    \hline
    \end{tabular}
    }
\end{table}

Furthermore, the performance of other loss function for the HSI PU learning task is also analyzed within the \textit{HOpenCls} framework, the proposed $\mathcal{L}_{tbce}$ get the best performance (Table~\ref{tab:diff_loss_results}).

\begin{table}[!t]
    \centering
    \caption{Results of different loss function of PU learning in the HOpenCls framework for unknown classes rejection. ↑ indicates larger values are better. ±$x$ denotes the standard error.}
    \label{tab:diff_loss_results}
    \resizebox{\linewidth}{!}{%
    \begin{tabular}{ccccccc} 
    \hline
    \multirow{2}{*}{$\mathcal{L}_{u}$}     & \multicolumn{2}{c}{WHU-Hi-HongHu}                                           & \multicolumn{2}{c}{WHU-Hi-LongKou}                                          & \multicolumn{2}{c}{University of  Pavia}                                     \\
                                       & Open OA↑                             & F1\textsuperscript{u}↑               & Open OA↑                             & F1\textsuperscript{u}↑               & Open OA↑                             & F1\textsuperscript{u}↑                \\ 
    \hline
    $\mathcal{L}_{Tar}$~\cite{T-HOneCls}      & 89.53\textsuperscript{±1.1}          & 71.80\textsuperscript{±2.1}          & 91.87\textsuperscript{±1.1}          & 90.59\textsuperscript{±1.2}          & 92.30\textsuperscript{±0.0}          & 74.02\textsuperscript{±0.0}           \\
    $\mathcal{L}_{tbce}^{w}$ & \textbf{96.03}\textsuperscript{±0.4} & \textbf{87.61}\textsuperscript{±1.2} & \textbf{99.15}\textsuperscript{±0.1} & \textbf{99.03}\textsuperscript{±0.1} & \textbf{97.17}\textsuperscript{±0.2} & \textbf{88.76}\textsuperscript{±0.7}  \\
    \hline
    \end{tabular}}
\end{table}

\subsection{Ablation Experiments Analysis}

Ablation experiments are conducted to evaluate the effectiveness of each module in the proposed \textit{HOpenCls}. The results are presented in Table~\ref{tab:ablation_experiments}. More detailed analysis can be found in the following:

\noindent \textbf{Multi-PU Head:}
The comparison between exp.1 and exp.2 highlights the significant role of the multi-PU head in the proposed \textit{HOpenCls}. For example, on the WHU-Hi-HongHu dataset, the Open OA, F1\textsuperscript{u}, and AUC\textsuperscript{u} scores improved by 27.66, 11.82, and 27.65, respectively. Similar improvements can also be observed in the comparison between exp.3 and exp.4.

\noindent \textbf{Grad-C Module:}
The comparison betweent exp.2 and exp.4 demonstrates that the Grad-C module is an effective solution for addressing the rejection of unknown classes in HSI. According to the Theorem~\ref{theorem}, the proposed $\mathcal{L}_{tbce}$ achieves better performance when the probability of the positive class is lower. Therefore, based on the combined experimental results of exp.1 to exp.4, greater improvements are observed when combining the multi-PU head with $\mathcal{L}_{tbce}$.

\noindent \textbf{Grad-E Module:}
Building upon $\mathcal{L}_{tbce}$, better open-set HSI classification results can be achieved by selectively restoring the gradient for wild unknown data, as demonstrated by the comparison between exp.4 and exp.5. Additionally, ensuring consistency between the two networks (exp.4 and exp.6) and applying the $\mathcal{L}_{tbce}^{w}$ (exp.6 and exp.7) further improves the performance of \textit{HOpenCls}.

\begin{table*}[!t]
    \centering
    \caption{Ablation experiments for different modules. ↑ indicates larger values are better. ±$x$ denotes the standard error.}
    \label{tab:ablation_experiments}
    \resizebox{\linewidth}{!}{%
    \begin{tabular}{ccccccccccccc} 
    \hline
    \multirow{2}{*}{Exp.} & \multicolumn{4}{c}{Ablation Experiments}                                                                                                  & \multicolumn{4}{c}{WHU-Hi-HongHu}                                                                                                                         & \multicolumn{4}{c}{WHU-Hi-LongKou}                                                                                                                         \\ 
    \cline{2-5}
                          & Grad-C Module\tablefootnote{A single network is employed in exp.1-exp.4.} & Grad-E Module                  & KL Loss      & Multi-PU Head & Open OA↑                             & F1\textsuperscript{u}↑               & AUC\textsuperscript{u}↑              & Close OA↑                            & Open OA↑                             & F1\textsuperscript{u}↑               & AUC\textsuperscript{u}↑              & Close OA↑                             \\ 
    \hline
    1                     & $\mathcal{L}_{bce}$                                                       &                                &              &               & 29.65\textsuperscript{±0.3}          & 26.74\textsuperscript{±0.1}          & 67.58\textsuperscript{±0.8}          & 95.01\textsuperscript{±0.8}          & 47.94\textsuperscript{±0.1}          & 59.88\textsuperscript{±0.1}          & 74.03\textsuperscript{±2.9}          & 99.48\textsuperscript{±0.1}           \\
    2                     & $\mathcal{L}_{bce}$                                                       &                                &              & $\checkmark$  & 57.31\textsuperscript{±11.1}         & 38.56\textsuperscript{±7.6}          & 95.23\textsuperscript{±2.9}          & 97.88\textsuperscript{±0.2}          & 68.78\textsuperscript{±1.6}          & 71.36\textsuperscript{±1.0}          & 98.59\textsuperscript{±0.5}          & 99.60\textsuperscript{±0.1}           \\
    3                     & $\mathcal{L}_{tbce}$                                                      &                                &              &               & 35.74\textsuperscript{±0.8}          & 28.95\textsuperscript{±0.3}          & 75.51\textsuperscript{±3.4}          & 92.63\textsuperscript{±1.3}          & 53.66\textsuperscript{±0.8}          & 62.64\textsuperscript{±0.4}          & 73.70\textsuperscript{±3.9}          & 98.99\textsuperscript{±0.5}           \\
    4                     & $\mathcal{L}_{tbce}$                                                      &                                &              & $\checkmark$  & 86.00\textsuperscript{±4.1}          & 66.40\textsuperscript{±7.6}          & 97.95\textsuperscript{±0.7}          & 97.34\textsuperscript{±0.6}          & 91.06\textsuperscript{±0.8}          & 89.76\textsuperscript{±0.8}          & 99.39\textsuperscript{±0.3}          & 99.44\textsuperscript{±0.1}           \\ 
    \cline{2-13}
    5                     & $\mathcal{L}_{tbce}$                                                      & $\mathcal{L}_{bce}^{w}$+MixPro &              & $\checkmark$  & 94.44\textsuperscript{±0.8}          & 83.90\textsuperscript{±2.6}          & 99.21\textsuperscript{±0.1}          & 98.15\textsuperscript{±0.6}          & 97.28\textsuperscript{±0.6}          & 96.74\textsuperscript{±0.7}          & 99.96\textsuperscript{±0.0}          & 99.53\textsuperscript{±0.1}           \\
    6                     & $\mathcal{L}_{tbce}$                                                      & $\mathcal{L}_{bce}^{w}$+MixPro & $\checkmark$ & $\checkmark$  & 94.59\textsuperscript{±0.5}          & 83.60\textsuperscript{±1.4}          & \textbf{99.28}\textsuperscript{±0.1} & 98.38\textsuperscript{±0.2}          & 97.92\textsuperscript{±0.1}          & 97.49\textsuperscript{±0.2}          & 99.98\textsuperscript{±0.0}          & 99.62\textsuperscript{±0.1}           \\
    7                     & $\mathcal{L}_{tbce}^{w}$+MixPro                                           & $\mathcal{L}_{bce}^{w}$+MixPro & $\checkmark$ & $\checkmark$  & \textbf{96.03}\textsuperscript{±0.4} & \textbf{87.61}\textsuperscript{±1.2} & 99.22\textsuperscript{±0.2}          & \textbf{98.57}\textsuperscript{±0.2} & \textbf{99.15}\textsuperscript{±0.1} & \textbf{99.03}\textsuperscript{±0.1} & \textbf{99.99}\textsuperscript{±0.0} & \textbf{99.75}\textsuperscript{±0.0}  \\
    \hline
    \end{tabular}
    }
    \end{table*}

\begin{figure*}[!t]
    \centering
    \subfloat[\small{Analysis of the order of Taylor series}]{
    \label{fig:taylor_series_analysis}
    \includegraphics[width=0.33\textwidth]{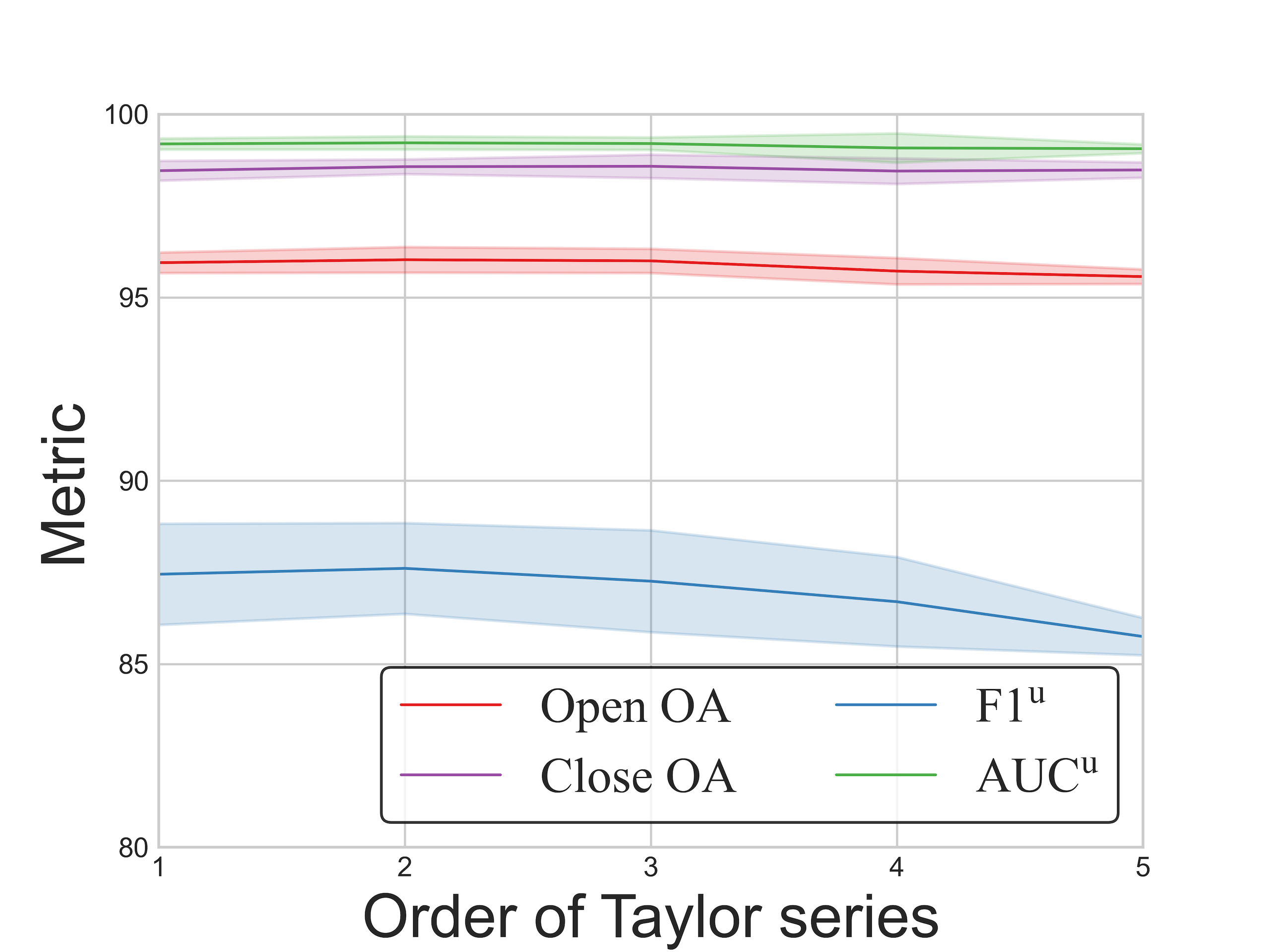}}
    \subfloat[\small{Analysis of the $\beta$}]{
    \label{fig:beta_analysis}
    \includegraphics[width=0.33\textwidth]{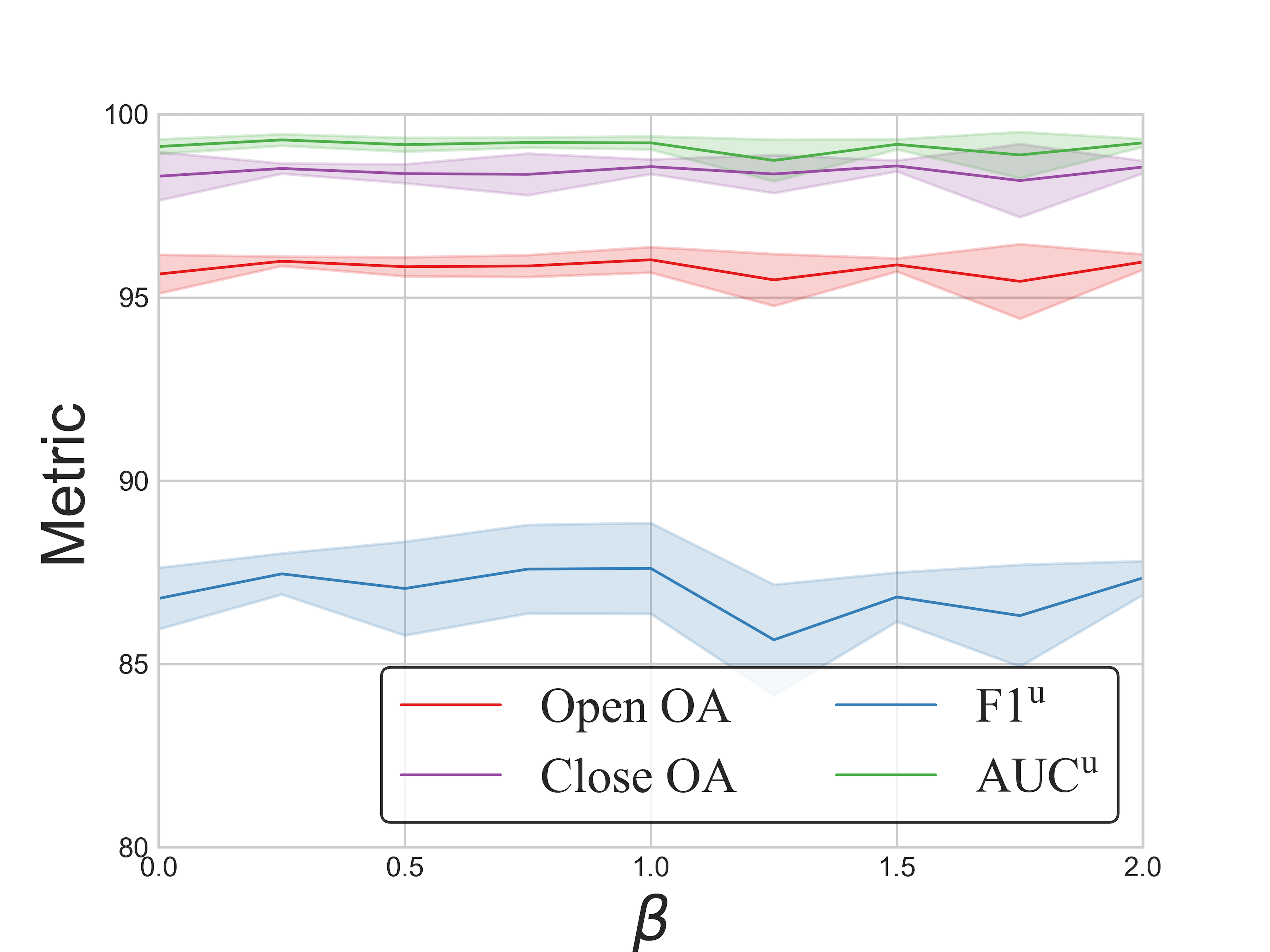}}
    \subfloat[\small{Analysis of the $\tau$}]{
    \label{fig:tao_analysis}
    \includegraphics[width=0.33\textwidth]{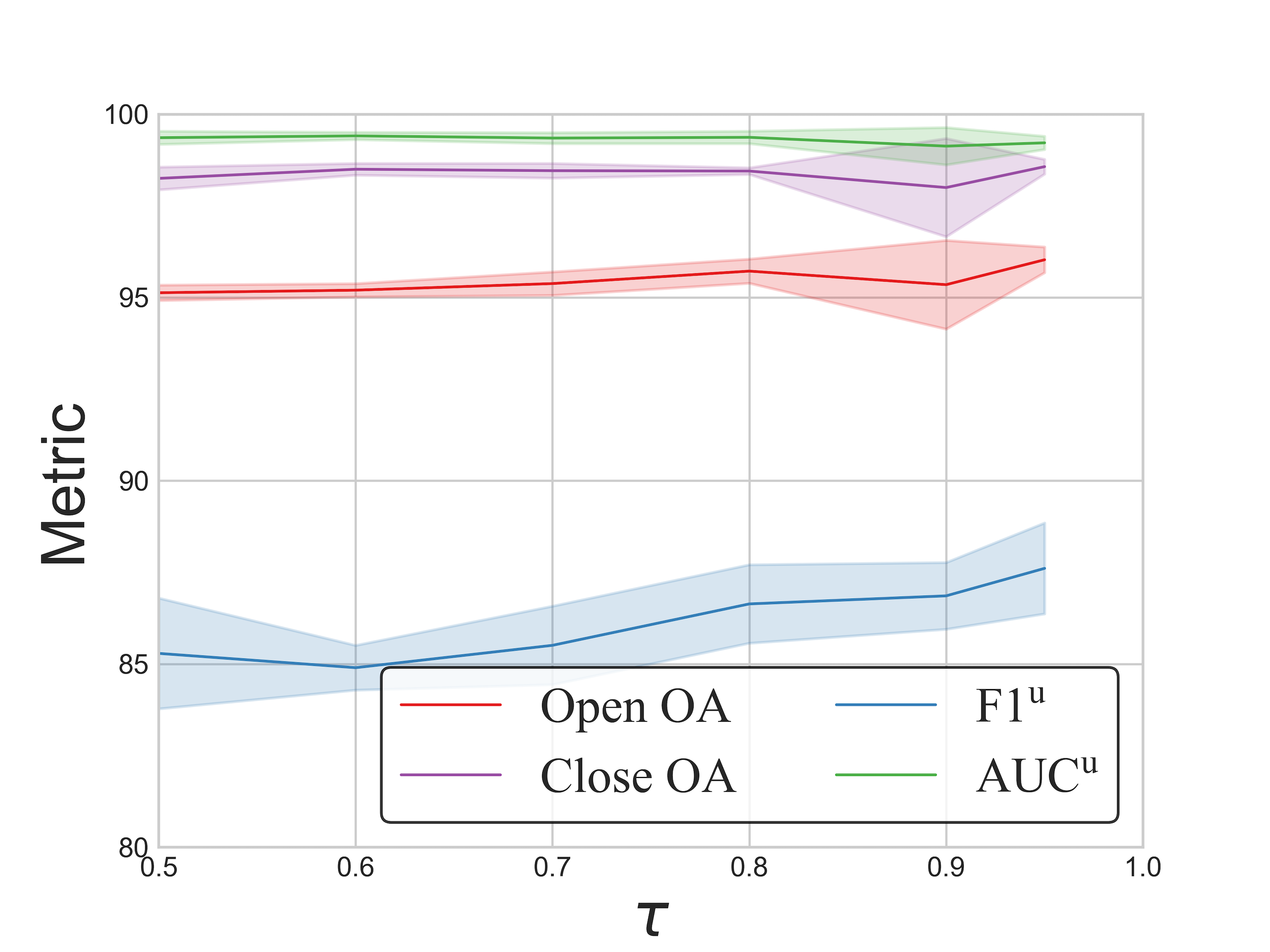}}
    \caption{Parametric Analysis of the HOpenCls framework.}
    \label{fig:parametric_analysis}
\end{figure*}

\noindent \textbf{Confidence Score Updating:}
Additional experiments were conducted to analyze the confidence scores updating strategies in the Grad-C and Grad-E modules, with results shown in Table~\ref{tab:confidence_update_stargeies}. The results demonstrate that discrete updating is more suited to the Grad-E module, while continuous updating works better for the Grad-C module. This suggests that less ambiguous information is more beneficial for the Grad-E module.

\begin{table}[!t]
    \centering
    \caption{Results of different confidence scores updating strategies. ↑ indicates larger values are better. ±$x$ denotes the standard error.}
    \label{tab:confidence_update_stargeies}
    \resizebox{\linewidth}{!}{%
    \begin{tabular}{cccccc} 
    \hline
    \multirow{2}{*}{$w_{c}$} & \multirow{2}{*}{$w_{e}$} & \multicolumn{2}{c}{WHU-Hi-HongHu}                                           & \multicolumn{2}{c}{WHU-Hi-LongKou}                                           \\
                             &                          & Open OA↑                             & F1\textsuperscript{u}↑               & Open OA↑                             & F1\textsuperscript{u}↑                \\ 
    \hline
    Continuous               & Continuous               & 94.43\textsuperscript{±0.5}          & 82.72\textsuperscript{±1.5}          & 97.03\textsuperscript{±0.3}          & 96.53\textsuperscript{±0.1}           \\
    Discrete                 & Discrete                 & 94.82\textsuperscript{±0.2}          & 81.46\textsuperscript{±1.1}          & 99.10\textsuperscript{±0.2}          & 98.96\textsuperscript{±0.3}           \\
    Discrete                 & Continuous               & 95.16\textsuperscript{±0.3}          & 84.64\textsuperscript{±1.0}          & 98.23\textsuperscript{±0.2}          & 97.83\textsuperscript{±0.2}           \\
    Continuous               & Discrete                 & \textbf{96.03}\textsuperscript{±0.4} & \textbf{87.61}\textsuperscript{±1.2} & \textbf{99.15}\textsuperscript{±0.1} & \textbf{99.03}\textsuperscript{±0.1}  \\
    \hline
    \end{tabular}
    }
\end{table}

\noindent \textbf{MixPro:}
Compared to the straightforward approach (Pro), MixPro leverages the consistency between the known classes classifier and multiple sub-PU classifiers. This strategy integrates the classification capability of the known classes classifier into the unknown classes rejection task, further enhancing the performance of the proposed \textit{HOpenCls}. (Table~\ref{tab:probability_mixture}).

\begin{table}[!t]
    \tiny
    \centering
    \caption{Analysis of the MixPro. ↑ indicates larger values are better. ±$x$ denotes the standard error.}
    \label{tab:probability_mixture}
    \resizebox{\linewidth}{!}{%
    \begin{tabular}{ccccc} 
    \hline
    \multirow{2}{*}{Pro/MixPro} & \multicolumn{2}{c}{WHU-Hi-HongHu}                                                             & \multicolumn{2}{c}{WHU-Hi-LongKou}                                                             \\
                                & Open OA↑                                      & F1\textsuperscript{u}↑                        & Open OA↑                                      & F1\textsuperscript{u}↑                         \\ 
    \hline
    Pro                         & 95.75\textsuperscript{±0.3}                   & 87.07\textsuperscript{±1.4}                   & 99.03\textsuperscript{±0.2}                   & 98.88\textsuperscript{±0.2}                    \\
    ProMix                      & \textbf{\textbf{96.03}}\textsuperscript{±0.4} & \textbf{\textbf{87.61}}\textsuperscript{±1.2} & \textbf{\textbf{99.15}}\textsuperscript{±0.1} & \textbf{\textbf{99.03}}\textsuperscript{±0.1}  \\
    \hline
    \end{tabular}
    }
\end{table}

\begin{table}[!t]
    \centering
    \caption{Results of different number of networks. ↑ indicates larger values are better. ±$x$ denotes the standard error.}
    \label{tab:number_of_networks}
    \resizebox{\linewidth}{!}{%
    \begin{tabular}{ccccc} 
    \hline
    \multirow{2}{*}{Number of Networks} & \multicolumn{2}{c}{WHU-Hi-HongHu}                                                             & \multicolumn{2}{c}{WHU-Hi-LongKou}                                                             \\
                                        & Open OA↑                                      & F1\textsuperscript{u}↑                        & Open OA↑                                      & F1\textsuperscript{u}↑                         \\ 
    \hline
    Single                              & 94.74\textsuperscript{±0.3}                   & 83.87\textsuperscript{±0.7}                   & 97.66\textsuperscript{±0.5}                   & 97.15\textsuperscript{±0.6}                    \\
    Two                                 & \textbf{\textbf{96.03}}\textsuperscript{±0.4} & \textbf{\textbf{87.61}}\textsuperscript{±1.2} & \textbf{\textbf{99.15}}\textsuperscript{±0.1} & \textbf{\textbf{99.03}}\textsuperscript{±0.1}  \\
    \hline
    \end{tabular}
    }
\end{table}

\noindent \textbf{Two Networks Cooperative Optimization:}
In the proposed \textit{HOpenCls}, two networks are optimized cooperatively to mitigate discrepancies (such as low-probability predictions) between $\mathcal{L}_{tbce}^{w}$ and $\mathcal{L}_{bce}^{w}$. As shown in Table~\ref{tab:number_of_networks}, optimizing a single network yields worse performance compared to the proposed \textit{HOpenCls}.

\noindent \textbf{Parametric Analysis}
Experiments were conducted to analyze the parametric sensitivity of \textit{HOpenCls}, with the results presented in Fig.~\ref{fig:parametric_analysis}. Several key observations can be made from the results: (1) The proposed \textit{HOpenCls} is robust to variations in $o$, $\beta$, and $\tau$; (2) Better unknown classes rejection results (F1\textsuperscript{u}) are achieved with lower-order Taylor series expansions, consistent with Theorem~\ref{theorem}; (3) Increasing $\tau$ leads to higher Open OA and F1\textsuperscript{u}, suggesting that the Grad-E module benefits from more accurate unknown classes data.

\section{Conclusion}
This paper proposes a novel open-set HSI classification framework utilizing wild data. Wild data holds significant promise due to its abundance, ease of collection, and better alignment with real-world data distributions. However, the mixed margin distribution of wild data, comprising both $\mathbb{P}_{k}$ and $\mathbb{P}_{u}$, presents challenges for its effective utilization. To overcome this challenge, this paper formulates it as a PU learning problem and specifically proposes a multi-PU head, Grad-C module, and Grad-E module to make it tractable. The results demonstrate that wild data can dramatically promote open-set HSI classification in practice, thereby helping to expedite the deployment of trustworthy models in complex real-world scenarios.


\appendix
\section*{Proof of the Theorem.\ref{theorem}}

Considering that the $\mathcal{L}_{tbce}(f(\boldsymbol{x}_{wild}),0)$ is bounded:
\begin{equation}
    0 \leq \mathcal{L}_{tbce}(f(\boldsymbol{x}_{wild}),0) \leq \mathcal{N}_{t},
\end{equation}
The effectiveness of the $\mathcal{L}_{tbce}$ for rejecting unknown classes can be proven as follows.

\begin{IEEEproof}[Proof of Theorem \ref{theorem}]
    From Eqn.~\ref{eq:huber_contamination_model}, we have
    \begin{equation}\nonumber
        \begin{aligned}
            {\mathcal{R}_{pu}(f)} &= \frac{1}{2}\left({\mathcal{R}_{k}^{+}(f)}+{\mathcal{R}_{wild}^{-}(f)}\right)\\
                                  &= \frac{1}{2}\left({\mathcal{R}_{k}^{+}(f)}+{\pi}{\mathcal{R}_{k}^{-}(f)}+{(1-\pi)}{\mathcal{R}_{u}^{-}(f)}\right)\\
                                  &= {R_{u}(f)}+\frac{1}{2}\left({\pi}{\mathcal{R}_{k}^{-}(f)}-{\pi}{\mathcal{R}_{u}^{-}(f)}\right),
        \end{aligned}
    \end{equation}
    where $\mathcal{R}^{-}_{k}(f)=\mathbb{E}_{(\boldsymbol{x}_{k},0){\sim}{\mathbb{P}_{k}}}\left[\mathcal{L}_{u}(f(\boldsymbol{x}),0)\right]$. Considered that $\mathcal{L}_{tbce}(f(\boldsymbol{x}_{wild}),0)$ is bounded:
    \begin{equation}\nonumber
        \begin{aligned}
            {\mathcal{R}_{pu}(f)} &\leq {\mathcal{R}_{u}(f)}+\frac{1}{2}{\pi}{\mathcal{R}_{k}^{-}(f)}\\
                                  &\leq {\mathcal{R}_{u}(f)}+\frac{1}{2}{\pi}{\mathcal{N}_t}
        \end{aligned}
    \end{equation}

    \begin{equation}\nonumber
        \begin{aligned}
            {\mathcal{R}_{pu}(f)} &\geq {\mathcal{R}_{u}(f)}-\frac{1}{2}{\pi}{\mathcal{R}_{u}^{-}(f)}\\
                                  &\geq {\mathcal{R}_{u}(f)}-\frac{1}{2}{{\pi}\mathcal{N}_{t}}.
        \end{aligned}
    \end{equation}
    Then we can ontain:
    \begin{equation}
        {\mathcal{R}_{pu}(f)}-{\frac{1}{2}\pi}{\mathcal{N}_t} \leq {\mathcal{R}_{u}(f)} \leq {\mathcal{R}_{pu}(f)}+{\frac{1}{2}{\pi}\mathcal{N}_{t}}
    \label{eq:bearing}
    \end{equation}
    The Theorem~\ref{theorem} can be proved as follows:
    \begin{equation}\nonumber
        \begin{aligned}
            {\mathcal{R}_{u}(\hat{f})}-{\mathcal{R}_{u}(f^{*})} \leq {\mathcal{R}_{pu}(\hat{f})}-{\mathcal{R}_{pu}(f^{*})}+{\pi}\mathcal{N}_{t} \leq {\pi}{\mathcal{N}_{t}}
        \end{aligned}
    \end{equation}
    Considered that $f^{*}$ is the global minimizers of the $\mathcal{R}_{u}(f)$:
    \begin{equation}\nonumber
        0 \leq {\mathcal{R}_{u}(\hat{f})}-{\mathcal{R}_{u}(f^{*})} \leq {{\pi}\mathcal{N}_{t}}.
    \end{equation}
    From Eqn.~\ref{eq:bearing}, we can obtain:
    \begin{equation}\nonumber
        {\mathcal{R}_{u}(f)}-{\frac{1}{2}{\pi}\mathcal{N}_{t}} \leq {\mathcal{R}_{pu}(f)} \leq {\mathcal{R}_{u}(f)}+{\frac{1}{2}{\pi}\mathcal{N}_{t}}.
    \end{equation}
    Then:
    \begin{equation}\nonumber
        \begin{aligned}
            {\mathcal{R}_{pu}(f^{*})}-{\mathcal{R}_{pu}(\hat{f})} \leq {\mathcal{R}_{u}(f^{*})}-{\mathcal{R}_{u}(\hat{f})}+{{\pi}\mathcal{N}_{t}} \leq {{\pi}\mathcal{N}_{t}}.
        \end{aligned}
    \end{equation}
    Considered that $\hat{f}$ is the global minimizers of the $\mathcal{R}_{pu}(f)$:
    \begin{equation}\nonumber
        0 \leq {\mathcal{R}_{pu}(f^{*})}-{\mathcal{R}_{pu}(\hat{f})} \leq {{\pi}\mathcal{N}_{t}}.
    \end{equation}
\end{IEEEproof}

{\small
\bibliographystyle{IEEEtran}
\bibliography{HOpenCls_ref}

\begin{thebibliography}{10}
\providecommand{\url}[1]{#1}
\csname url@samestyle\endcsname
\providecommand{\newblock}{\relax}
\providecommand{\bibinfo}[2]{#2}
\providecommand{\BIBentrySTDinterwordspacing}{\spaceskip=0pt\relax}
\providecommand{\BIBentryALTinterwordstretchfactor}{4}
\providecommand{\BIBentryALTinterwordspacing}{\spaceskip=\fontdimen2\font plus
\BIBentryALTinterwordstretchfactor\fontdimen3\font minus \fontdimen4\font\relax}
\providecommand{\BIBforeignlanguage}[2]{{%
\expandafter\ifx\csname l@#1\endcsname\relax
\typeout{** WARNING: IEEEtran.bst: No hyphenation pattern has been}%
\typeout{** loaded for the language `#1'. Using the pattern for}%
\typeout{** the default language instead.}%
\else
\language=\csname l@#1\endcsname
\fi
#2}}
\providecommand{\BIBdecl}{\relax}
\BIBdecl

\bibitem{6555921}
J.~M. Bioucas-Dias, A.~Plaza, G.~Camps-Valls, P.~Scheunders, N.~Nasrabadi, and J.~Chanussot, ``Hyperspectral remote sensing data analysis and future challenges,'' \emph{IEEE Geoscience and Remote Sensing Magazine}, vol.~1, no.~2, pp. 6--36, 2013.

\bibitem{FPGA}
Z.~Zheng, Y.~Zhong, A.~Ma, and L.~Zhang, ``Fpga: Fast patch-free global learning framework for fully end-to-end hyperspectral image classification,'' \emph{IEEE Transactions on Geoscience and Remote Sensing}, vol.~58, no.~8, pp. 5612--5626, 2020.

\bibitem{10078841}
X.~Ou, M.~Wu, B.~Tu, G.~Zhang, and W.~Li, ``Multi-objective unsupervised band selection method for hyperspectral images classification,'' \emph{IEEE Transactions on Image Processing}, vol.~32, pp. 1952--1965, 2023.

\bibitem{10696913}
J.~Li, Z.~Zhang, Y.~Liu, R.~Song, Y.~Li, and Q.~Du, ``Swformer: Stochastic windows convolutional transformer for hybrid modality hyperspectral classification,'' \emph{IEEE Transactions on Image Processing}, pp. 1--1, 2024.

\bibitem{10167502}
C.~Zhao, B.~Qin, S.~Feng, W.~Zhu, W.~Sun, W.~Li, and X.~Jia, ``Hyperspectral image classification with multi-attention transformer and adaptive superpixel segmentation-based active learning,'' \emph{IEEE Transactions on Image Processing}, vol.~32, pp. 3606--3621, 2023.

\bibitem{WHU-Hi}
Y.~Zhong, X.~Hu, C.~Luo, X.~Wang, J.~Zhao, and L.~Zhang, ``Whu-hi: Uav-borne hyperspectral with high spatial resolution (h2) benchmark datasets and classifier for precise crop identification based on deep convolutional neural network with crf,'' \emph{Remote Sensing of Environment}, vol. 250, p. 112012, 2020.

\bibitem{ITreeDet}
H.~Zhao, Y.~Zhong, X.~Wang, X.~Hu, C.~Luo, M.~Boitt, R.~Piiroinen, L.~Zhang, J.~Heiskanen, and P.~Pellikka, ``Mapping the distribution of invasive tree species using deep one-class classification in the tropical montane landscape of kenya,'' \emph{ISPRS Journal of Photogrammetry and Remote Sensing}, vol. 187, pp. 328--344, 2022.

\bibitem{WANG2022113058}
J.~Wang, A.~Ma, Y.~Zhong, Z.~Zheng, and L.~Zhang, ``Cross-sensor domain adaptation for high spatial resolution urban land-cover mapping: From airborne to spaceborne imagery,'' \emph{Remote Sensing of Environment}, vol. 277, p. 113058, 2022.

\bibitem{10325566}
J.~Yang, B.~Du, D.~Wang, and L.~Zhang, ``Iter: Image-to-pixel representation for weakly supervised hsi classification,'' \emph{IEEE Transactions on Image Processing}, vol.~33, pp. 257--272, 2024.

\bibitem{10047983}
J.~Yang, B.~Du, Y.~Xu, and L.~Zhang, ``Can spectral information work while extracting spatial distribution?—an online spectral information compensation network for hsi classification,'' \emph{IEEE Transactions on Image Processing}, vol.~32, pp. 2360--2373, 2023.

\bibitem{9573256}
Y.~Xu, B.~Du, and L.~Zhang, ``Self-attention context network: Addressing the threat of adversarial attacks for hyperspectral image classification,'' \emph{IEEE Transactions on Image Processing}, vol.~30, pp. 8671--8685, 2021.

\bibitem{MDL4OW}
S.~Liu, Q.~Shi, and L.~Zhang, ``Few-shot hyperspectral image classification with unknown classes using multitask deep learning,'' \emph{IEEE Transactions on Geoscience and Remote Sensing}, vol.~59, no.~6, pp. 5085--5102, 2021.

\bibitem{Fang_OpenSet}
J.~Yue, L.~Fang, and M.~He, ``Spectral-spatial latent reconstruction for open-set hyperspectral image classification,'' \emph{IEEE Transactions on Image Processing}, vol.~31, pp. 5227--5241, 2022.

\bibitem{Kang_OpenSet}
Z.~Xie, P.~Duan, W.~Liu, X.~Kang, X.~Wei, and S.~Li, ``Feature consistency-based prototype network for open-set hyperspectral image classification,'' \emph{IEEE Transactions on Neural Networks and Learning Systems}, vol.~35, no.~7, pp. 9286--9296, 2024.

\bibitem{OpenMax}
A.~Bendale and T.~E. Boult, ``Towards open set deep networks,'' in \emph{2016 IEEE Conference on Computer Vision and Pattern Recognition (CVPR)}, 2016, pp. 1563--1572.

\bibitem{Entropy}
R.~Chan, M.~Rottmann, and H.~Gottschalk, ``Entropy maximization and meta classification for out-of-distribution detection in semantic segmentation,'' in \emph{2021 IEEE/CVF International Conference on Computer Vision (ICCV)}, 2021, pp. 5108--5117.

\bibitem{WOODS}
J.~Katz-Samuels, J.~B. Nakhleh, R.~Nowak, and Y.~Li, ``Training {OOD} detectors in their natural habitats,'' in \emph{Proceedings of the 39th International Conference on Machine Learning}, ser. Proceedings of Machine Learning Research, K.~Chaudhuri, S.~Jegelka, L.~Song, C.~Szepesvari, G.~Niu, and S.~Sabato, Eds., vol. 162.\hskip 1em plus 0.5em minus 0.4em\relax PMLR, 17--23 Jul 2022, pp. 10\,848--10\,865.

\bibitem{Energy}
W.~Liu, X.~Wang, J.~Owens, and Y.~Li, ``Energy-based out-of-distribution detection,'' in \emph{Advances in Neural Information Processing Systems}, H.~Larochelle, M.~Ranzato, R.~Hadsell, M.~Balcan, and H.~Lin, Eds., vol.~33.\hskip 1em plus 0.5em minus 0.4em\relax Curran Associates, Inc., 2020, pp. 21\,464--21\,475.

\bibitem{Huber}
P.~J. Huber, \emph{Robust Estimation of a Location Parameter}.\hskip 1em plus 0.5em minus 0.4em\relax New York, NY: Springer New York, 1992, pp. 492--518.

\bibitem{DistPU}
Y.~Jiang, Q.~Xu, Y.~Zhao, Z.~Yang, P.~Wen, X.~Cao, and Q.~Huang, ``Positive-unlabeled learning with label distribution alignment,'' \emph{IEEE Transactions on Pattern Analysis and Machine Intelligence}, vol.~45, no.~12, pp. 15\,345--15\,363, 2023.

\bibitem{T-HOneCls}
H.~Zhao, X.~Wang, J.~Li, and Y.~Zhong, ``Class prior-free positive-unlabeled learning with taylor variational loss for hyperspectral remote sensing imagery,'' in \emph{2023 IEEE/CVF International Conference on Computer Vision (ICCV)}, 2023, pp. 16\,781--16\,790.

\bibitem{nnPU}
R.~Kiryo, G.~Niu, M.~C. du~Plessis, and M.~Sugiyama, ``Positive-unlabeled learning with non-negative risk estimator,'' in \emph{Advances in Neural Information Processing Systems}, I.~Guyon, U.~V. Luxburg, S.~Bengio, H.~Wallach, R.~Fergus, S.~Vishwanathan, and R.~Garnett, Eds., vol.~30.\hskip 1em plus 0.5em minus 0.4em\relax Curran Associates, Inc., 2017.

\bibitem{PUET}
J.~Wilton, A.~Koay, R.~Ko, M.~Xu, and N.~Ye, ``Positive-unlabeled learning using random forests via recursive greedy risk minimization,'' in \emph{Advances in Neural Information Processing Systems}, S.~Koyejo, S.~Mohamed, A.~Agarwal, D.~Belgrave, K.~Cho, and A.~Oh, Eds., vol.~35.\hskip 1em plus 0.5em minus 0.4em\relax Curran Associates, Inc., 2022, pp. 24\,060--24\,071.

\bibitem{HOneCls}
H.~Zhao, Y.~Zhong, X.~Wang, and H.~Shu, ``One-class risk estimation for one-class hyperspectral image classification,'' \emph{IEEE Transactions on Geoscience and Remote Sensing}, vol.~61, pp. 1--17, 2023.

\bibitem{10400402}
J.~Li, Z.~Zhang, R.~Song, Y.~Li, and Q.~Du, ``Scformer: Spectral coordinate transformer for cross-domain few-shot hyperspectral image classification,'' \emph{IEEE Transactions on Image Processing}, vol.~33, pp. 840--855, 2024.

\bibitem{10050427}
Y.~Zhang, W.~Li, W.~Sun, R.~Tao, and Q.~Du, ``Single-source domain expansion network for cross-scene hyperspectral image classification,'' \emph{IEEE Transactions on Image Processing}, vol.~32, pp. 1498--1512, 2023.

\bibitem{9785505}
X.~Zheng, H.~Sun, X.~Lu, and W.~Xie, ``Rotation-invariant attention network for hyperspectral image classification,'' \emph{IEEE Transactions on Image Processing}, vol.~31, pp. 4251--4265, 2022.

\bibitem{8737729}
Y.~Xu, B.~Du, and L.~Zhang, ``Beyond the patchwise classification: Spectral-spatial fully convolutional networks for hyperspectral image classification,'' \emph{IEEE Transactions on Big Data}, vol.~6, no.~3, pp. 492--506, 2020.

\bibitem{HU2022147}
X.~Hu, X.~Wang, Y.~Zhong, and L.~Zhang, ``S3anet: Spectral-spatial-scale attention network for end-to-end precise crop classification based on uav-borne h2 imagery,'' \emph{ISPRS Journal of Photogrammetry and Remote Sensing}, vol. 183, pp. 147--163, 2022.

\bibitem{9347487}
D.~Wang, B.~Du, and L.~Zhang, ``Fully contextual network for hyperspectral scene parsing,'' \emph{IEEE Transactions on Geoscience and Remote Sensing}, vol.~60, pp. 1--16, 2022.

\bibitem{9040673}
C.~Geng, S.-J. Huang, and S.~Chen, ``Recent advances in open set recognition: A survey,'' \emph{IEEE Transactions on Pattern Analysis and Machine Intelligence}, vol.~43, no.~10, pp. 3614--3631, 2021.

\bibitem{9857485}
K.~Kirchheim, M.~Filax, and F.~Ortmeier, ``Pytorch-ood: A library for out-of-distribution detection based on pytorch,'' in \emph{2022 IEEE/CVF Conference on Computer Vision and Pattern Recognition Workshops (CVPRW)}, 2022, pp. 4350--4359.

\bibitem{MSP}
D.~Hendrycks and K.~Gimpel, ``A baseline for detecting misclassified and out-of-distribution examples in neural networks,'' in \emph{International Conference on Learning Representations}, 2017.

\bibitem{ODIN}
S.~Liang, Y.~Li, and R.~Srikant, ``Enhancing the reliability of out-of-distribution image detection in neural networks,'' in \emph{International Conference on Learning Representations}, 2018.

\bibitem{KLMatching}
D.~Hendrycks, S.~Basart, M.~Mazeika, A.~Zou, J.~Kwon, M.~Mostajabi, J.~Steinhardt, and D.~Song, ``Scaling out-of-distribution detection for real-world settings,'' in \emph{Proceedings of the 39th International Conference on Machine Learning}, vol. 162, 2022, pp. 8759--8773.

\bibitem{8953952}
R.~Yoshihashi, W.~Shao, R.~Kawakami, S.~You, M.~Iida, and T.~Naemura, ``Classification-reconstruction learning for open-set recognition,'' in \emph{2019 IEEE/CVF Conference on Computer Vision and Pattern Recognition (CVPR)}, 2019, pp. 4011--4020.

\bibitem{9296325}
H.-M. Yang, X.-Y. Zhang, F.~Yin, Q.~Yang, and C.-L. Liu, ``Convolutional prototype network for open set recognition,'' \emph{IEEE Transactions on Pattern Analysis and Machine Intelligence}, vol.~44, no.~5, pp. 2358--2370, 2022.

\bibitem{CACLoss}
D.~Miller, N.~Sünderhauf, M.~Milford, and F.~Dayoub, ``Class anchor clustering: A loss for distance-based open set recognition,'' in \emph{2021 IEEE Winter Conference on Applications of Computer Vision (WACV)}, 2021, pp. 3569--3577.

\bibitem{10415443}
Y.~Du, X.~Li, L.~Shi, F.~Li, and T.~Xu, ``A prototype network for hyperspectral image open-set classification based on feature invariance and weighted pearson distance measurement,'' \emph{IEEE Transactions on Geoscience and Remote Sensing}, vol.~62, pp. 1--17, 2024.

\bibitem{OE}
A.~R. Dhamija, M.~G\"{u}nther, and T.~Boult, ``Reducing network agnostophobia,'' in \emph{Advances in Neural Information Processing Systems}, vol.~31, 2018.

\bibitem{EOS}
D.~Hendrycks, M.~Mazeika, and T.~Dietterich, ``Deep anomaly detection with outlier exposure,'' in \emph{International Conference on Learning Representations}, 2019.

\bibitem{FOODY20061}
G.~M. Foody, A.~Mathur, C.~Sanchez-Hernandez, and D.~S. Boyd, ``Training set size requirements for the classification of a specific class,'' \emph{Remote Sensing of Environment}, vol. 104, no.~1, pp. 1--14, 2006.

\bibitem{Gong_Wang_Ye_Xu_Lin_2018}
T.~Gong, G.~Wang, J.~Ye, Z.~Xu, and M.~Lin, ``Margin based pu learning,'' \emph{Proceedings of the AAAI Conference on Artificial Intelligence}, vol.~32, no.~1, Apr. 2018.

\bibitem{9201373}
W.~Li, Q.~Guo, and C.~Elkan, ``One-class remote sensing classification from positive and unlabeled background data,'' \emph{IEEE Journal of Selected Topics in Applied Earth Observations and Remote Sensing}, vol.~14, pp. 730--746, 2021.

\bibitem{LU2021112584}
Y.~Lu and L.~Wang, ``How to automate timely large-scale mangrove mapping with remote sensing,'' \emph{Remote Sensing of Environment}, vol. 264, p. 112584, 2021.

\bibitem{ijcai2019p590}
C.~Zhang, D.~Ren, T.~Liu, J.~Yang, and C.~Gong, ``Positive and unlabeled learning with label disambiguation,'' in \emph{Proceedings of the Twenty-Eighth International Joint Conference on Artificial Intelligence, {IJCAI-19}}, 7 2019, pp. 4250--4256.

\bibitem{kato2018learning}
M.~Kato, T.~Teshima, and J.~Honda, ``Learning from positive and unlabeled data with a selection bias,'' in \emph{International Conference on Learning Representations}, 2019.

\bibitem{LI2022102947}
J.~Li, X.~Wang, H.~Zhao, X.~Hu, and Y.~Zhong, ``Detecting pine wilt disease at the pixel level from high spatial and spectral resolution uav-borne imagery in complex forest landscapes using deep one-class classification,'' \emph{International Journal of Applied Earth Observation and Geoinformation}, vol. 112, p. 102947, 2022.

\bibitem{PAN}
W.~Hu, R.~Le, B.~Liu, F.~Ji, J.~Ma, D.~Zhao, and R.~Yan, ``Predictive adversarial learning from positive and unlabeled data,'' \emph{Proceedings of the AAAI Conference on Artificial Intelligence}, vol.~35, no.~9, pp. 7806--7814, May 2021.

\bibitem{vPU}
H.~Chen, F.~Liu, Y.~Wang, L.~Zhao, and H.~Wu, ``A variational approach for learning from positive and unlabeled data,'' in \emph{Advances in Neural Information Processing Systems}, vol.~33, 2020, pp. 14\,844--14\,854.

\bibitem{p3mix}
C.~Li, X.~Li, L.~Feng, and J.~Ouyang, ``Who is your right mixup partner in positive and unlabeled learning,'' in \emph{International Conference on Learning Representations}, 2022.

\bibitem{DS3L}
L.-Z. Guo, Z.-Y. Zhang, Y.~Jiang, Y.-F. Li, and Z.-H. Zhou, ``Safe deep semi-supervised learning for unseen-class unlabeled data,'' in \emph{Proceedings of the 37th International Conference on Machine Learning}, vol. 119, 2020, pp. 3897--3906.

\end{thebibliography}
}

\end{document}